\documentclass[runningheads]{llncs}

\usepackage[year=2026]{eccv}

\usepackage{eccvabbrv}

\usepackage{graphicx}
\usepackage{booktabs}
\usepackage{multirow}

\usepackage[accsupp]{axessibility}

\usepackage{hyperref}

\usepackage{orcidlink}
\usepackage{amsmath}
\usepackage{amsfonts}\usepackage{amsmath}
\usepackage{amsfonts}
\usepackage{amssymb}
\usepackage{enumitem}

\newcommand\blfootnote[1]{\begingroup\renewcommand\thefootnote{}\footnotetext{#1}\endgroup}

\begin{document}

\title{Resonant Minds: Closed-Loop Social Avatars with Theory of Mind} 

\titlerunning{Resonant Minds}

\author{Jianxu Shangguan\inst{1}\textsuperscript{*}, Jing Xu\inst{2}\textsuperscript{*}, Hang Ye\inst{2}, Xiaoxuan Ma\inst{3}, \\Yizhou Wang\inst{2}, Jenq-Neng Hwang\inst{1}, Wentao Zhu\inst{4}}

\authorrunning{J.~Shangguan et al.}

\institute{$^{1}$University of Washington \qquad $^{2}$Peking University \\
$^{3}$Carnegie Mellon University \qquad $^{4}$Eastern Institute of Technology, Ningbo}

\maketitle
\blfootnote{\textsuperscript{*}\,Equal contribution.}

\begin{abstract}
  Creating lifelike digital humans with genuine social intelligence requires unifying cognitive reasoning and multimodal generation within a coherent framework. Current approaches treat these as separate tasks: Large Language Models excel at dialogue but lack embodied expression, while diffusion-based talking head models achieve visual fidelity but ignore social cognition. To bridge this gap, we propose a closed-loop dual-agent framework integrating perception, social reasoning, and expression into a continuous interaction cycle. The perception module analyzes partners' multimodal behaviors from video, while the social reasoning module infers hidden mental states through Theory of Mind and selects responses via an ensemble mechanism. The expression module then generates emotion-controllable videos that jointly synthesize speaker speech and facial expressions with listener reactive behaviors, capturing bidirectional dynamics absent in prior work. We further construct a hierarchical Persona-Scenario dataset with psychologically grounded personas and private social goals to support evaluation under information asymmetry. Experiments on this dataset demonstrate competitive or superior performance on both dialogue quality and video generation metrics. Notably, our method surpasses even the full-information Script mode on key dialogue quality dimensions, suggesting that explicit mental state inference under uncertainty can elicit more thoughtful dialogue than unrestricted information access. Project page: \url{https://resonantminds.github.io/}.
\end{abstract}

\vspace{-3pt}
\section{Introduction}
\label{sec:intro}

\begin{figure*}[!t]
        \centering
        \includegraphics[width=\textwidth]{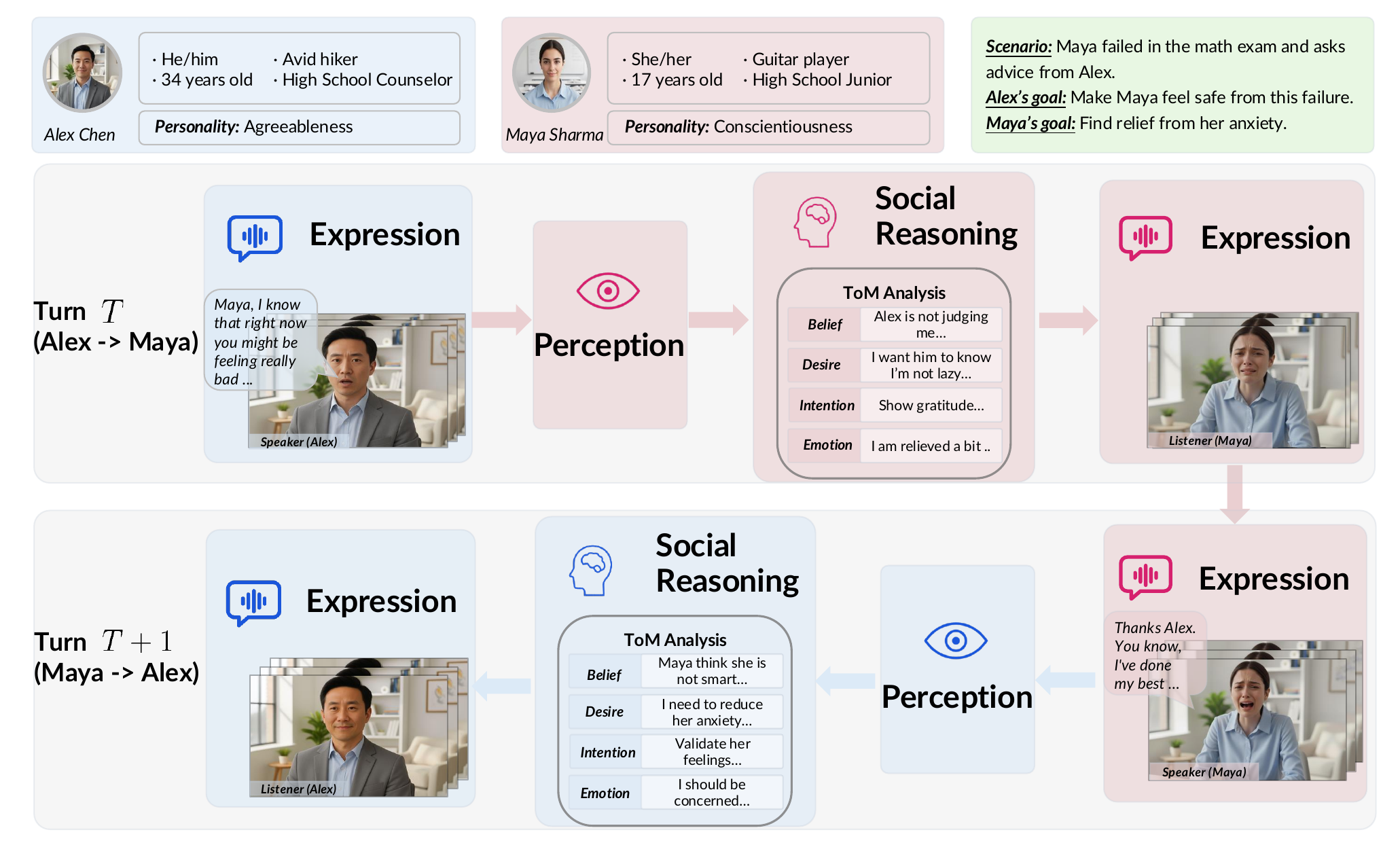}
        \vspace{-2ex}
        \caption{
            \textbf{Overview of our dual-agent interaction framework.} Given a conversational scenario where each agent possesses a private persona profile hidden from their partner, our framework generates strategic, emotionally-aware multimodal responses through three integrated modules. At turn $T$, Agent \textit{Maya} first perceives Agent \textit{Alex}'s video through the perception module. The social reasoning module then performs Theory of Mind (ToM) analysis to deduce \textit{Alex}'s hidden mental state, generates candidate responses, and employs an ensemble mechanism to select the optimal action balancing empathy, strategy, and personality coherence. The expression module synthesizes this action into a multimodal video for \textit{Alex} at turn $T+1$, completing the closed-loop interaction.
        }
        \label{fig:teaser}
        \vspace{-4ex}
\end{figure*}

Creating lifelike digital human agents capable of engaging in nuanced and meaningful social interactions is a grand challenge in artificial intelligence. The applications range from collaborative dialogue partners~\cite{he2017learning,sotopia} to embodied agents navigating complex social scenarios~\cite{kovavc2021socialai, li2023camel, lowe2017multi,sotopia}, with concrete downstream use cases including emotionally responsive game NPCs~\cite{agentgame}, virtual social-skills training~\cite{agent_social_train}, and simulated counseling companions~\cite{counsellor_client}. These applications require not just fluent text or realistic videos but genuine social intelligence~\cite{Cognitive_psychology}. This challenge arises because human social interaction is a remarkably complex process, extending far beyond the literal exchange of words~\cite{kihlstrom2000social,tomasello2019becoming}. The complexity stems from a layered cognitive foundation. Individuals possess personality traits~\cite{de2002big,lang2011short} that consistently shape their behavioral patterns, while successful interaction requires first perceiving each other, Theory of Mind (ToM) reasoning~\cite{premack1978does} to infer their hidden traits and transient mental states from observable behaviors. Thus, an ideal social agent must therefore holistically integrate strategic social reasoning under information asymmetry, emotionally-aware multimodal perception and expression, and consistent personality-driven behavior.

This pursuit, however, is hindered by a structural disconnection between high-level social reasoning and low-level multimodal generation. While Large Language Models (LLM) have demonstrated human-like behaviors on cognitive evaluations~\cite{achiam2023gpt,team2023gemini,liu2024deepseek,aher2023using,eval-big-five}, they still falter in realistic social simulations under information asymmetry (\emph{Agent mode}), where agents must infer hidden mental states, rather than operating with full access to all agents' private information (\emph{Script mode})~\cite{script_vs_agent}. Simultaneously, diffusion-based talking head models, despite excelling at visual fidelity, focus predominantly on single-person speaking~\cite{dice-talk,ji2025sonic,xu2025hunyuanportrait,lin2025mvportrait}, one-way listening~\cite{inproceedings2}, or open-loop dyadic switching~\cite{inproceedings}, neglecting the closed-loop dual-agent interaction fundamental to real conversations. Exacerbating these disconnections, existing approaches further overlook the foundational cognitive mechanisms of ToM reasoning and personality-driven behavior essential for realistic social interaction.

In light of this, we propose a novel closed-loop multimodal dual-agent interaction framework inspired by cognitive insights. Each agent is endowed with a rich persona profile, including background, personality traits, and private social goals that remain hidden from their conversation partner. In turn-by-turn interactions within shared scenarios, agents must perceive and infer the other agent's hidden mental states from observable behaviors while pursuing their own objectives. This design enables us to investigate how stable personality systematically shapes emotional expression and reasoning across sustained social interactions.
Built upon this setting, our method operates through a Perception-Social Reasoning-Expression framework with three tightly integrated modules. As shown in Fig.~\ref{fig:teaser}, given two persona profiles and scenario as input, our system runs a closed-loop cycle where agents dynamically exchange speaker and listener roles: 1) Perception module analyzes the partner's audio prosody, emotion state, and facial expressions from the previous turn; 2) Social Reasoning module employs a ToM analysis to infer the partner's hidden mental states and an ensemble mechanism to select responses balancing empathy, strategy, and coherence; 3) Expression module translates this response into synchronized speech and video with fine-grained emotional control, completing the interaction loop. Existing persona datasets contain either shallow trait descriptions or lack private goals necessary for studying meaningful interactions~\cite{faithful-persona,lee-etal-2022-personachatgen,zhang-etal-2018-personalizing}. To support evaluation, we construct a Persona-Scenario dataset with psychologically grounded personas based on the Big Five framework~\cite{kwantes2016assessing, lang2011short}. Personas are built through a sparse-to-rich pipeline that generates detailed personality narratives from factual backgrounds, then situated in diverse scenarios adapted from Sotopia~\cite{sotopia}, where agents pursue private goals under information asymmetry.

By situating agents in shared scenarios with persistent personas and private goals, our framework enables simultaneous optimization of social interaction quality and multimodal generation fidelity. Experiments demonstrate competitive or superior performance on both dialogue quality and video generation metrics. Notably, our method surpasses even the full-information Script mode on key dialogue quality dimensions, suggesting that explicit mental state inference under uncertainty can elicit more thoughtful dialogue than unrestricted information access. Overall, our contributions are summarized as follows:
\begin{itemize}
\item We formulate dual-agent multimodal conversation under information asymmetry, where personality-driven agents infer hidden mental states while generating emotionally expressive responses.
\item We propose a closed-loop dual-agent interaction framework integrating perception, social reasoning, and expression, producing personality consistent and emotionally expressive multimodal interactions.
\item We construct a hierarchical Persona-Scenario dataset with psychologically grounded personas, achieving competitive or superior performance on both dialogue generation and video synthesis in comprehensive experiments.
\end{itemize}

\vspace{-3pt}
\section{Related Work}
\label{sec:related_work}

\vspace{-2pt}
\subsection{LLM-based Social Simulation}
Recent research has leveraged LLMs to simulate human social behavior~\cite{aher2023using,Character-llm,sotopia,eval-big-five,script_vs_agent}. A critical limitation is the neglect of information asymmetry~\cite{script_vs_agent,Oey,tomasello2009cultural}: script mode grants a single LLM full access to all agents' private goals~\cite{li2023camel,interactive,Symmetric_Collaborative_Dialogue}, leading to unrealistic strategies such as information leakage~\cite{script_vs_agent,Symmetric_Collaborative_Dialogue}, whereas agent mode restricts each LLM to its own persona---yet existing text-only agent-mode systems lack multimodal behavioral cues essential for social inference~\cite{sotopia,shuster2021iyoustateoftheartdialogue,sharma2025understandingsycophancylanguagemodels}. A parallel line of work investigates computational Theory of Mind (ToM). Classical Bayesian models~\cite{baker2017rational} are principled but computationally intractable in open-ended natural language settings; even recent methods such as AutoToM~\cite{zhang2025autotom} rely on LLM backends. Empirical studies show that GPT-4-series models achieve human-level ToM performance on false-belief tasks, indirect requests, and misdirection~\cite{strachan2024testing}, whereas no prior social simulation work integrates LLM-based ToM with multimodal generation. Our work bridges both gaps: we operate under realistic agent-mode information asymmetry, and equip each agent with an LLM-based ToM module that infers the partner's hidden mental states from multimodal behavioral cues to drive emotionally-controllable dual-agent video synthesis.

\vspace{-2pt}
\subsection{Talking Head Generation}

Recent advances in talking head generation have focused on improving audio-visual synchronization and emotional expressiveness~\cite{dice-talk,ji2025sonic,tan2025edtalk}. Methods like DICE-Talk~\cite{dice-talk} leverage emotion embeddings for controllable generation, while Sonic~\cite{ji2025sonic} emphasizes global audio perception for enhanced realism. EDTalk~\cite{tan2025edtalk} introduces disentangled representations to separately control emotions and identity. However, these approaches share a fundamental limitation: they generate only single-character talking heads, lacking the capability to model conversational dynamics between multiple participants. Recent works explore interactive scenarios: ViCo~\cite{zhou2022responsive} focuses on listener non-verbal reactions; Learning2Listen~\cite{ng2022learninglistenmodelingnondeterministic} and MFR-Net~\cite{liu2023mfrnetmultifacetedresponsivelistening} model listener responses with limited emotion control. INFP~\cite{inproceedings} generates dyadic conversational video by switching each agent between speaker and listener states, yet its bidirectionality is open-loop, driven by audio alone without language, persona, or across-turn feedback. Across these methods the interaction lacks a cognitive feedback loop, so a perceived cue never shapes the partner's reasoning on the following turn. In contrast, our framework is the first to implement closed-loop bidirectional interaction in which both agents alternate between speaking and listening roles. Each agent's perceived multimodal cues update an explicit ToM state that conditions its response across multi-turn dialogue. Furthermore, existing methods treat emotion control as static input rather than dynamic conversational outcome, lacking mechanisms to infer appropriate emotions from dialogue context. Our work integrates dynamic listener generation with LLM-based social reasoning, enabling conversational videos where both agents exhibit contextually appropriate emotional expressions.

\section{Method}
\label{sec:method}

\begin{figure*}[!t]
    \centering
    \includegraphics[width=\linewidth]{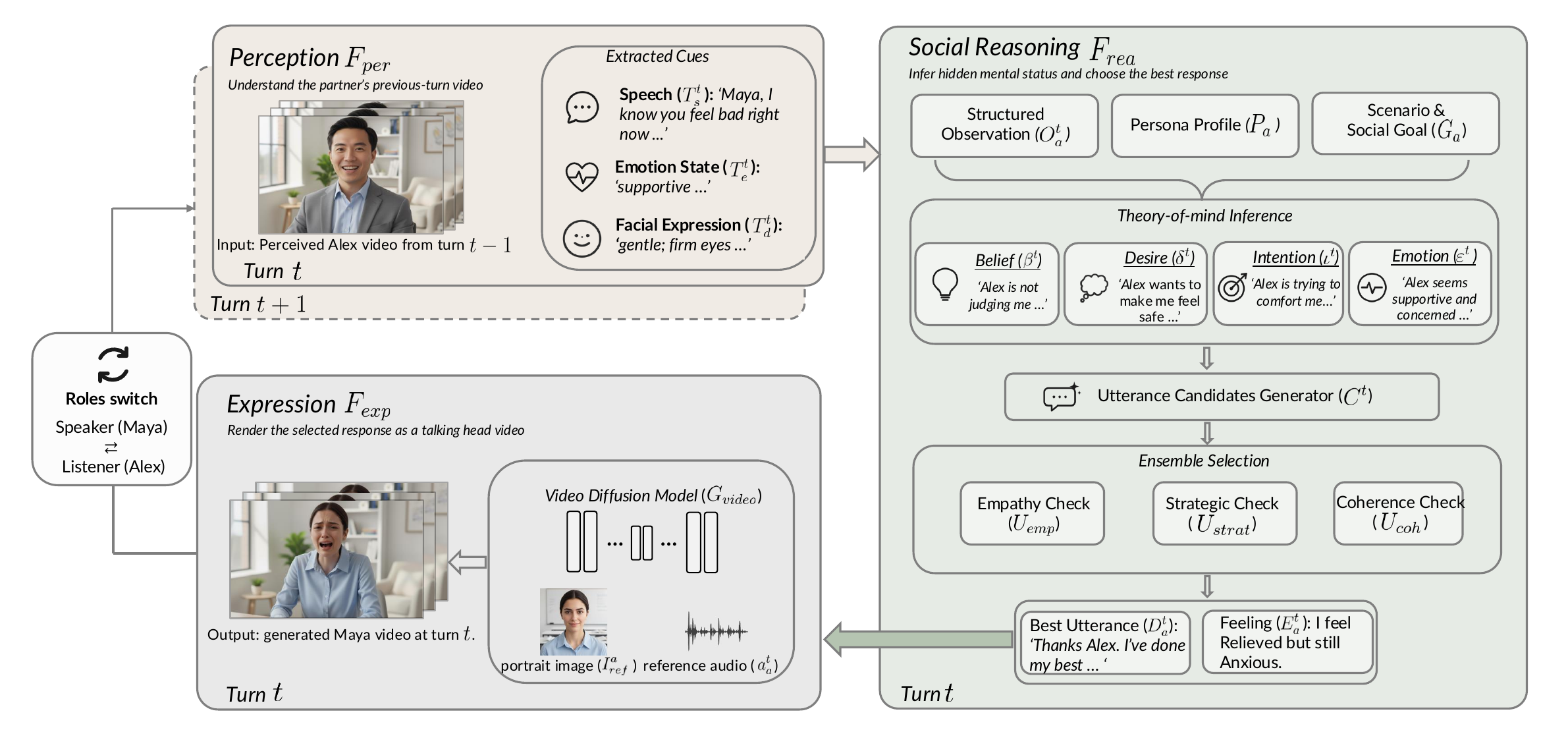}
    \vspace{-4ex}
    \caption{
    \textbf{Overview of our closed-loop framework for dual-agent interaction.} Given two agents, each holding a private persona profile and sharing the same scenario as in Fig.~\ref{fig:teaser}, our goal is to generate strategic, emotionally-aware responses under information asymmetry (Sec.~\ref{sec:method_overview}). At turn $t$, the framework comprises three modules: 1) \textbf{Perception} ($\mathcal{F}_{\text{per}}$) extracts structured observations $o_\mathfrak{a}^t$ from the partner's previous-turn video (Sec.~\ref{sec:perception}). 2) \textbf{Social Reasoning} ($\mathcal{F}_{\text{rea}}$) employs ToM analysis and an ensemble mechanism to select the optimal action $A_\mathfrak{a}^t$ (Sec.~\ref{sec:social_reasoning}). 3) \textbf{Expression} ($\mathcal{F}_{\text{exp}}$) renders the selected action into the agent's multimodal response video through emotionally-controllable dual-agent synthesis (Sec.~\ref{sec:expression}). At turn $t{+}1$, the two agents switch their speaker and listener roles, and the newly generated video drives the next perception step, closing the interaction loop.
}
    \vspace{-3ex}
\label{fig:pipeline}
\end{figure*}

\subsection{Overview}
\label{sec:method_overview}

High-level social reasoning under uncertainty and low-level emotional multimodal expression are interdependent challenges that necessitate joint modeling through a holistic framework.

We model dual-agent social interaction as a Partially Observable Markov Game (POMG) defined by $\langle \mathcal{S}, \mathcal{A}, \mathcal{O}, \mathcal{T} \rangle$, where $\mathcal{S}$ represents world states, $\mathcal{A} = \mathcal{A}_\mathfrak{a} \times \mathcal{A}_\mathfrak{b}$ denotes the joint action space, $\mathcal{O} = \mathcal{O}_\mathfrak{a} \times \mathcal{O}_\mathfrak{b}$ is the joint observation space, and $\mathcal{T}$ defines state transitions. Each agent $\mathfrak{i} \in \{\mathfrak{a}, \mathfrak{b}\}$ is characterized by a persistent persona profile $P_\mathfrak{i} = (\mathcal{B}_\mathfrak{i}, \Psi_\mathfrak{i}, \mathcal{G}_\mathfrak{i})$, where $\mathcal{B}_\mathfrak{i}$ is a biographical background narrative, $\Psi_\mathfrak{i}$ is a rich personality narrative grounded in the Big Five dimensions (Openness, Conscientiousness, Extraversion, Agreeableness, Neuroticism) that describes behavioral patterns and interaction styles, and $\mathcal{G}_\mathfrak{i}$ specifies private social goals invisible to the partner. All components are stored as structured JSON format. At each turn $t$, agent $\mathfrak{i}$ receives observation $o_\mathfrak{i}^t \in \mathcal{O}_\mathfrak{i}$ capturing the partner's observable behaviors, then selects action $A_\mathfrak{i}^t \in \mathcal{A}_\mathfrak{i}$ to advance its goals while maintaining persona consistency.

Our framework implements a policy $\pi_{\theta_\mathfrak{i}}$ that maps observations to strategic actions through three integrated modules operating in a closed loop, as illustrated in Fig.~\ref{fig:pipeline}. At each conversational turn $t$, the agent first employs the \textbf{Perception module}(Sec.\ref{sec:perception}) ($\mathcal{F}_{\text{per}}$) to analyze the partner's video from the previous turn $V_\mathfrak{i}^{t-1}$, extracting structured textual observations that capture speech content, emotional state, and facial expressions. These observations are then fed into the \textbf{Social Reasoning module} (Sec.\ref{sec:social_reasoning})($\mathcal{F}_{\text{rea}}$), which performs ToM inference based on the BDIE framework to deduce the partner's hidden mental states—beliefs, desires, intentions, and emotions. The module generates multiple candidate responses and evaluates them through an ensemble mechanism comprising three specialized evaluators assessing empathy, strategy, and coherence, ultimately selecting the optimal action $A_\mathfrak{i}^t = (D_\mathfrak{i}^t, E_\mathfrak{i}^t)$ that best balances these objectives. This symbolic action, consisting of a textual utterance and emotional descriptor, is passed to the \textbf{Expression module} (Sec.\ref{sec:expression})($\mathcal{F}_{\text{exp}}$), which translates it into a multimodal video $V_\mathfrak{i}^t$ through a text-to-emotion adapter and dual-agent video synthesis. The generated video becomes the partner's observable input at turn $t+1$, completing the closed-loop cycle where agents dynamically exchange speaker-listener roles. We detail each module in the following subsections.

\subsection{Multimodal Perception Module}
\label{sec:perception}
The Multimodal Perception module $\mathcal{F}_{\text{per}}$ transforms the partner's video stream into structured textual observations for downstream reasoning. Given the partner's video $V_\mathfrak{b}^{t-1}$ consisting of visual frames $v_\mathfrak{b}^{t-1} \in \mathbb{R}^{T \times H \times W \times 3}$ and audio $a_\mathfrak{b}^{t-1} \in \mathbb{R}^{T_a}$, the module generates:
\begin{equation}
  o_\mathfrak{a}^t = \mathcal{F}_{\text{per}}(V_\mathfrak{b}^{t-1}) = (T_s^t, T_e^t, T_d^t),
  \label{eq:perception}
\end{equation}
where $T_s^t$ is the transcribed speech, $T_e^t$ is the inferred emotional state, and $T_d^t$ describes facial expressions. We employ HumanOmni~\cite{humanomni} with first-person prompting to extract these components from Agent $\mathfrak{a}$'s observational perspective, preserving the information asymmetry of agent mode.

\subsection{Social Reasoning Module}
\label{sec:social_reasoning}

The Social Reasoning module $\mathcal{F}_{\text{rea}}$ constitutes the cognitive core of our framework, transforming multimodal observations into strategic, persona-consistent responses. Under information asymmetry, the module first infers the partner's hidden mental states through ToM analysis, then employs an ensemble mechanism to select actions that balance emotional appropriateness, strategic goal advancement, and personality coherence. Formally, the module maps the current observation $o_\mathfrak{a}^t$, interaction history $\mathcal{H}_\mathfrak{a}^{t-1}$, and persona profile $P_\mathfrak{a}$ to an updated history $\mathcal{H}_\mathfrak{a}^t$ and the selected action $A_\mathfrak{a}^t$:
\begin{equation}
  \mathcal{H}_\mathfrak{a}^t, A_\mathfrak{a}^t = \mathcal{F}_{\text{rea}}(o_\mathfrak{a}^t, \mathcal{H}_\mathfrak{a}^{t-1}, P_\mathfrak{a}).
  \label{eq:social_reasoning}
\end{equation}
All three phases---ToM Analysis, Candidate Generation, and Ensemble Mechanism are implemented via structured GPT-4o prompting with chain-of-thought elicitation.

\subsubsection{Theory of Mind Analysis}
\label{sec:tom}
The ToM module component $\mathcal{F}_{\text{ToM}}$ performs structured mental state attribution, inferring the partner's hidden cognitive states from observable behaviors.
We adopt the BDIE framework from cognitive psychology to structure the mental state representation. Given the latest observation $o_\mathfrak{a}^t = (T_s^t, T_e^t, T_d^t)$, interaction history $\mathcal{H}_\mathfrak{a}^{t-1}$, and scenario context $\xi$, the ToM inference generates an estimated mental state attribution:
\begin{equation}
  \Phi_{\text{ToM}}^t = \hat{M}_\mathfrak{b}^t = \mathcal{F}_{\text{ToM}}(o_\mathfrak{a}^t, \mathcal{H}_\mathfrak{a}^{t-1}, \xi) = (\beta^t, \delta^t, \iota^t, \epsilon^t),
  \label{eq:tom}
\end{equation}
where each component captures a distinct facet of the partner's inferred mental state: \textbf{Belief ($\beta^t$):} What Agent $\mathfrak{b}$ understands about the current situation, including their interpretation of Agent $\mathfrak{a}$'s previous actions. \textbf{Desire ($\delta^t$):} Agent $\mathfrak{b}$'s underlying needs, preferences, and intrinsic motivations. \textbf{Intention ($\iota^t$):} Agent $\mathfrak{b}$'s tactical objectives and strategic agenda in the current turn. \textbf{Emotion ($\epsilon^t$):} Agent $\mathfrak{b}$'s current affective state and the appropriate emotional response from Agent $\mathfrak{a}$.
All four facets are represented as free-form natural language text, enabling flexible and context-sensitive mental state description.
\subsubsection{Candidate Response Generation}
Before committing to an action, humans subconsciously generate multiple candidate replies that differ in tone, strategy, and emotional framing. Mimicking this process, our framework employs a candidate generation strategy $\mathcal{G}_{\text{cand}}$ that produces $K$ diverse options for downstream deliberation. Given the ToM module analysis $\Phi_{\text{ToM}}^t$, persona profile $P_\mathfrak{a}$, and interaction history $\mathcal{H}_\mathfrak{a}^{t-1}$, the generator produces $K$ candidate actions:
\begin{equation}
  \mathcal{C}^t = \{A_{\mathfrak{a},1}^t, \ldots, A_{\mathfrak{a},K}^t\} = \mathcal{G}_{\text{cand}}(\Phi_{\text{ToM}}^t, P_\mathfrak{a}, \mathcal{H}_\mathfrak{a}^{t-1}),
  \label{eq:candidate}
\end{equation}
where each candidate $A_{\mathfrak{a},i}^t = (D_{\mathfrak{a},i}^t, E_{\mathfrak{a},i}^t)$ comprises a textual utterance and an emotional expression descriptor.

\subsubsection{Ensemble Mechanism}
\label{sec:ensemble}
The final action selection employs an ensemble mechanism that evaluates each candidate through multiple specialized criteria, mirroring how humans weigh competing considerations before committing to an action. The ensemble comprises three specialized evaluators: \textbf{Empathy Evaluator ($\mathcal{E}_{\text{emp}}$):} Assesses whether the candidate appropriately acknowledges and responds to the partner's observed emotional state. \textbf{Strategy Evaluator ($\mathcal{E}_{\text{strat}}$):} Evaluates the candidate's strategic value for advancing the agent's private goal. \textbf{Coherence Evaluator ($\mathcal{E}_{\text{coh}}$):} Measures consistency with the agent's established persona and dialogue history.
Each evaluator $\mathcal{E}_j$ is implemented as a GPT-4o judge that assigns an integer score on a 0--10 scale, which is then normalized to $U_j(A_{\mathfrak{a},i}^t) \in [0,1]$. The final selection aggregates these evaluations through weighted voting:
\begin{equation}
  A_\mathfrak{a}^t = \operatorname*{arg\,max}_{A_{\mathfrak{a},i}^t \in \mathcal{C}^t} \sum_{j} \lambda_j U_j(A_{\mathfrak{a},i}^t),
  \label{eq:ensemble_selection}
\end{equation}
where $j \in \{\text{emp}, \text{strat}, \text{coh}\}$ denotes empathy, strategic, and coherence evaluators respectively. We set $\lambda_{\text{emp}} = \lambda_{\text{strat}} = \lambda_{\text{coh}} = 1/3$ as the default equal weighting; a sensitivity analysis over different $\lambda$ is provided in Supplementary~\ref{sec:experiment_details}.

\subsection{Expression Module}
\label{sec:expression}

The Expression module $\mathcal{F}_{\text{exp}}$ translates the symbolic action $A_\mathfrak{a}^t = (D_\mathfrak{a}^t, E_\mathfrak{a}^t)$ into a multimodal video stream:
\begin{equation}
  V_\mathfrak{a}^t = \mathcal{F}_{\text{exp}}(A_\mathfrak{a}^t, P_\mathfrak{a}, I_{\text{ref}}^\mathfrak{a}),
  \label{eq:expression}
\end{equation}
where $I_{\text{ref}}^\mathfrak{a}$ is the agent's reference portrait image.

\subsubsection{Text-to-Emotion Adapter}

A critical challenge is bridging the semantic-to-parametric gap: the Social Reasoning module produces natural language emotion descriptors $E_\mathfrak{a}^t$, while talking head models require numerical representations. We introduce a Text-to-Weights Adapter $\mathcal{A}_{\phi}$ that maps textual emotions to DICE-Talk's~\cite{dice-talk} 8-dimensional emotion weight space:
\begin{equation}
  \mathbf{w}_\mathfrak{a}^t = \mathcal{A}_{\phi}(\text{Enc}_{\text{CLIP}}(E_\mathfrak{a}^t)),
  \label{eq:adapter}
\end{equation}
where $\text{Enc}_{\text{CLIP}}$ is a pretrained CLIP text encoder~\cite{clip}, and $\mathcal{A}_{\phi}$ is a learned MLP mapping semantic embeddings to normalized weight vectors ($\sum_i w_i = 1$, $w_i \geq 0$).

\subsubsection{Dual-Agent Video Synthesis}
The module synthesizes emotionally-aware speech via IndexTTS~\cite{zhou2025indextts2breakthroughemotionallyexpressive}: $a_\mathfrak{a}^t = \mathcal{F}_{\text{TTS}}(D_\mathfrak{a}^t, E_\mathfrak{a}^t)$, then generates the speaker video using DICE-Talk~\cite{dice-talk} conditioned on audio $a_\mathfrak{a}^t$, emotion weights $\mathbf{w}_\mathfrak{a}^t$, and identity $I_{\text{ref}}^\mathfrak{a}$:
\begin{equation}
  v_\mathfrak{a}^t = \mathcal{G}_{\text{video}}(a_\mathfrak{a}^t, \mathbf{w}_\mathfrak{a}^t, I_{\text{ref}}^\mathfrak{a}),
  \label{eq:video_gen}
\end{equation}

The dual-agent composition mechanism renders both the speaker and the reactive listener into a single video:
\begin{equation}
  V_{\text{full}}^t = \mathcal{F}_{\text{compose}}(v_\mathfrak{a}^t, v_\mathfrak{b}^t),
  \label{eq:composition}
\end{equation}
where the listener video $v_\mathfrak{b}^t$ is synthesized symmetrically to the speaker. The listener's personality $\Psi_\mathfrak{b}$ and the perceived speaker action $A_\mathfrak{a}^t$ condition a reactive emotion descriptor $E_\mathfrak{b}^t$, which the same Text-to-Emotion Adapter $\mathcal{A}_{\phi}$ maps to an 8-dimensional weight vector $\mathbf{w}_\mathfrak{b}^t$ that drives DICE-Talk to render $v_\mathfrak{b}^t$. This personality-driven reactive modeling enables bidirectional nonverbal communication, with the listener's micro-expressions conveying empathetic social signals. The operator $\mathcal{F}_{\text{compose}}$ then synthesizes the two streams in parallel, concatenates them horizontally into a $1024\times512$ frame, applies RIFE interpolation~\cite{huang2022realtimeintermediateflowestimation} to raise the frame rate from $12.5$ to $25$ FPS, and merges the dual-TTS audio tracks. The generated video $V_\mathfrak{a}^t$ becomes the partner's observation at turn $t+1$, completing the closed loop.

\begin{figure*}[!t]
  \centering
  \begin{subfigure}{0.30\textwidth}
      \includegraphics[width=\linewidth]{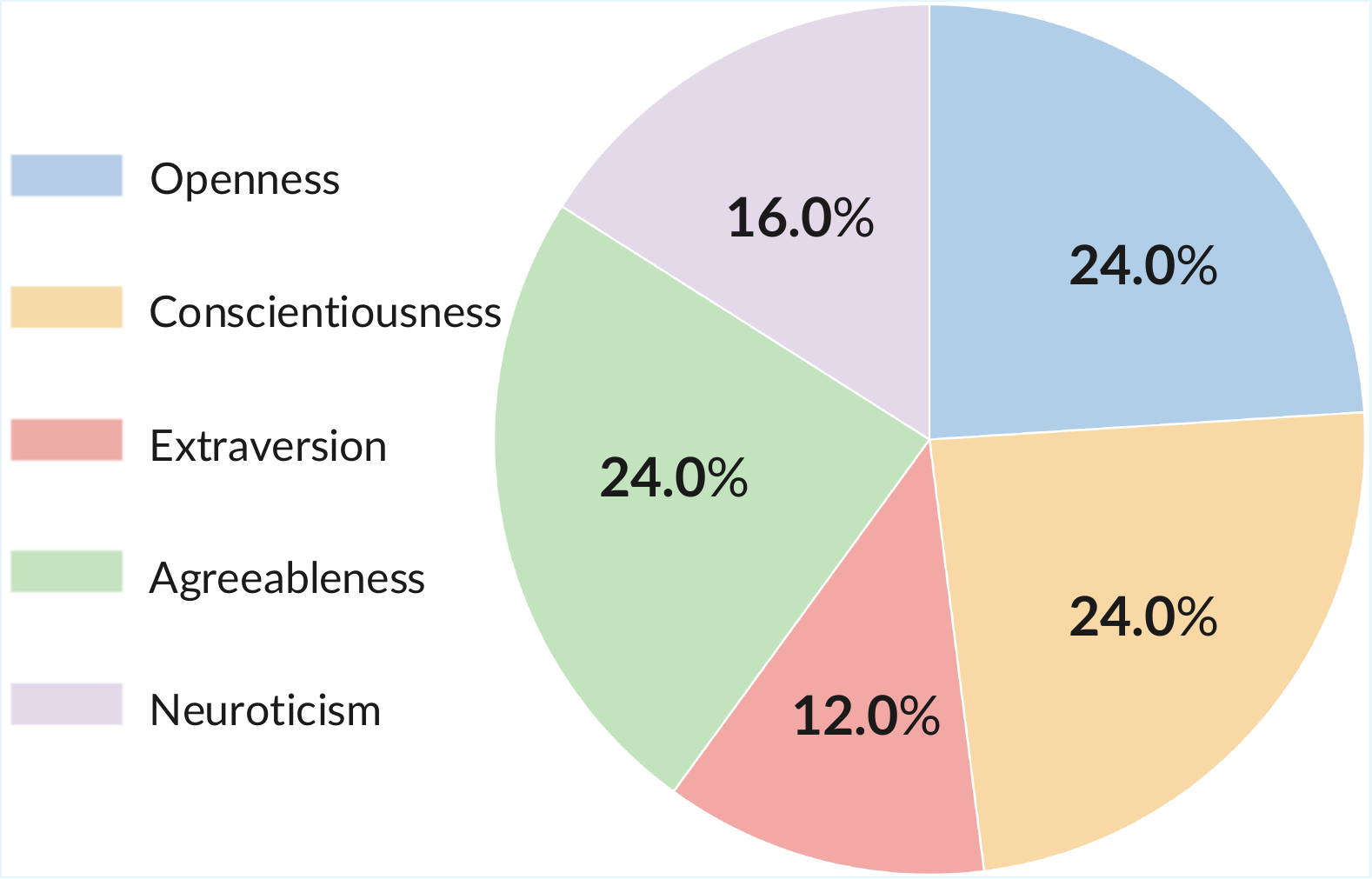} 
      \caption{Big Five Personality Distribution}
      \label{fig:dataset_a}
  \end{subfigure}
  \hfill
  \begin{subfigure}{0.30\textwidth}
      \includegraphics[width=\linewidth]{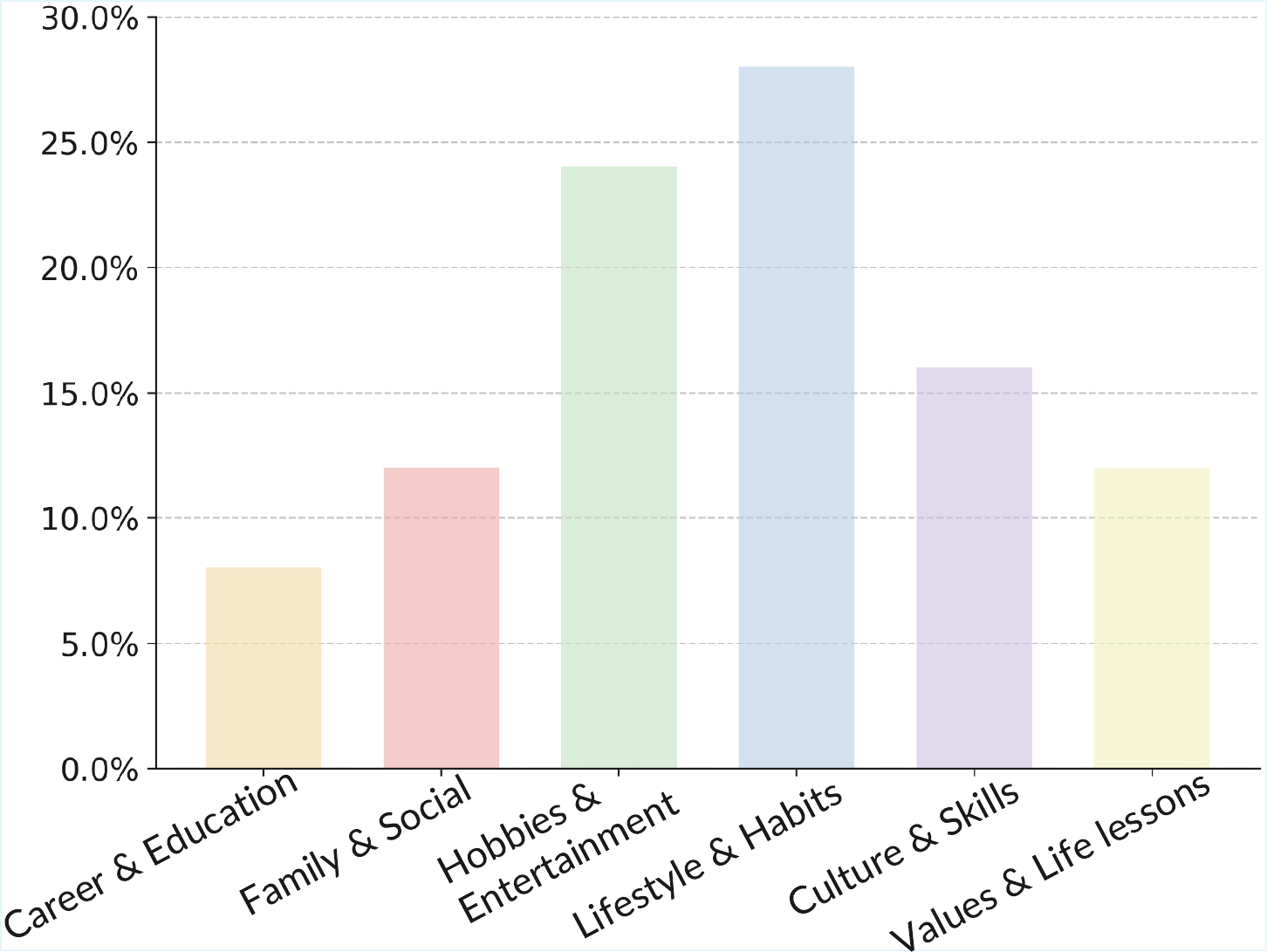}
      \caption{Attributes Distribution}
      \label{fig:dataset_b}
  \end{subfigure}
  \hfill
  \begin{subfigure}{0.30\textwidth}
      \includegraphics[width=\linewidth]{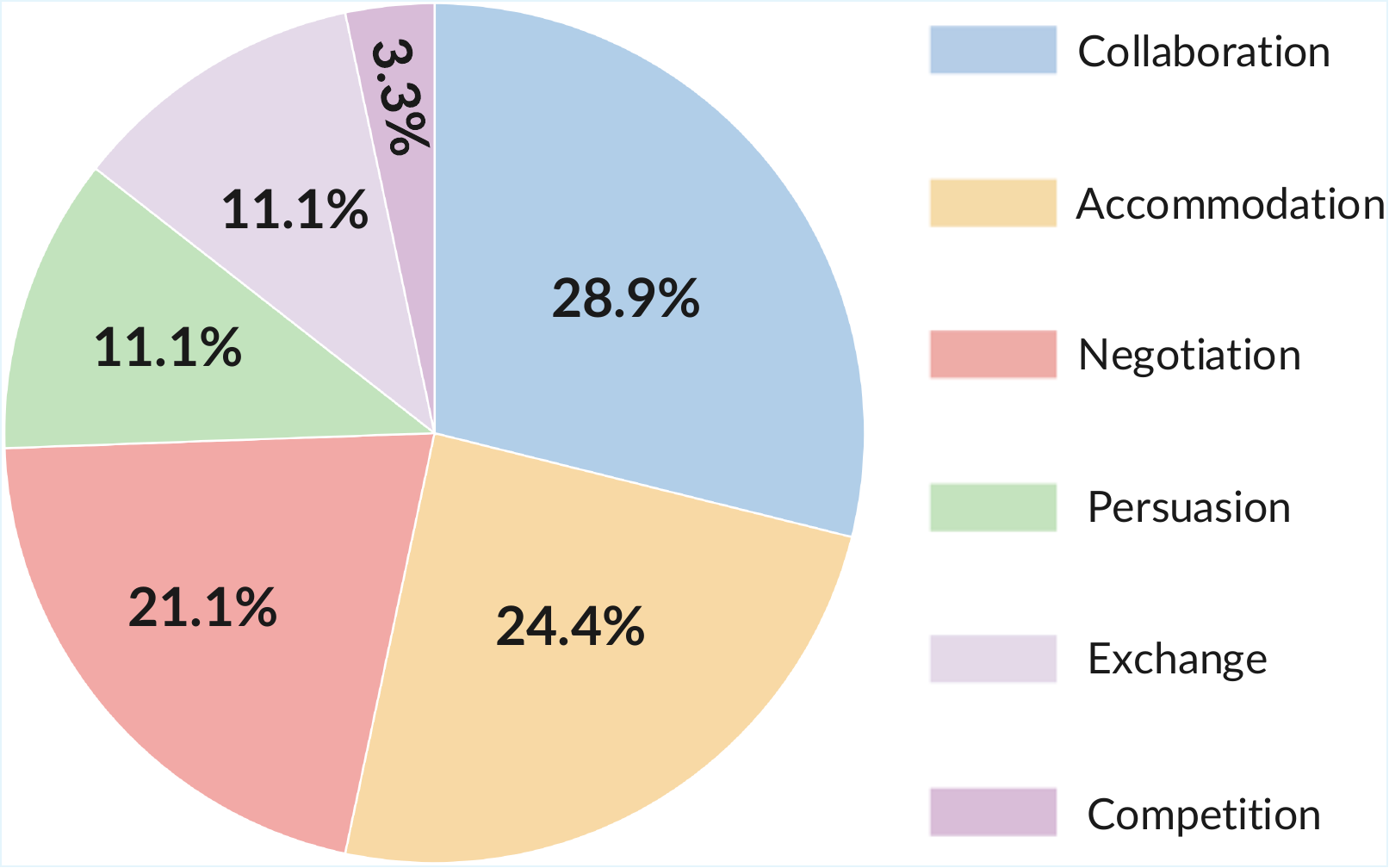}
      \caption{Scenario Distribution}
      \label{fig:dataset_c}
  \end{subfigure}
  \vspace{-1ex}
  \caption{\textbf{Persona-Scenario dataset statistics.} The dataset contains 50 personas generated via GPT-4o with NLI consistency checking, each comprising a facts layer (biographical attributes) and a psychological layer (Big Five personality narratives), paired with 90 social scenarios from Sotopia~\cite{sotopia}. (a)~Big Five personality scores are approximately uniformly distributed to avoid trait bias. (b)~Factual attributes span diverse demographic categories. (c)~Scenarios cover six interaction types (e.g., negotiation, collaboration, persuasion) to evaluate social reasoning under varied conditions.}
  \label{fig:dataset}
\end{figure*}

\subsection{Persona-Scenario Dataset}

\label{sec:dataset}
To support evaluation under information asymmetry, we construct a hierarchical Persona-Scenario dataset using GPT-4o generation with human verification, following a Sparse-to-Rich pipeline. Each of our 50 personas comprises two layers: a facts layer containing logically consistent biographical attributes and a Psychological Layer with rich behavioral narratives grounded in the Big Five framework~\cite{de2002big,eval-big-five}. The personas exhibit balanced distribution across personality dimensions and are situated in 90 diverse social scenarios from Sotopia~\cite{sotopia}, spanning six interaction types including negotiation, collaboration, and persuasion. For example, a facts layer entry may specify \textit{``34-year-old female architect based in Chicago, divorced, passionate about sustainable design.''} while the corresponding psychological layer describes behavioral patterns such as \textit{``high Openness and low Neuroticism; she approaches conflicts with curiosity rather than defensiveness and readily explores unconventional solutions.''}
Figure~\ref{fig:dataset} illustrates the personality distribution, background attributes diversity and scenario coverage, with implementation details and sample cases in Supplementary~\ref{sec:dataset_creation}.

\vspace{-3pt}
\section{Experiments}
\label{sec:experiment}

\vspace{-2pt}
\subsection{Experiment Settings}
\label{subsec:implementation}

\textbf{System Configuration.}
The system uses GPT-4o for dialogue generation. Speaker audio is synthesized using Index-TTS v2~\cite{zhou2025indextts2breakthroughemotionallyexpressive} with emotion control, combined with ElevenLabs for listener backchannel generation. Video generation is implemented using DICE-Talk~\cite{dice-talk} with emotion control enabled. Final videos are rendered side-by-side with RIFE interpolation~\cite{huang2022realtimeintermediateflowestimation} for temporal smoothing. Full implementation details are provided in Supplementary~\ref{sec:implementation_details}.

\noindent\textbf{Dialogue Baselines.}
We compare against two dialogue generation baselines operating under different information access paradigms.
\textit{Agent mode}~\cite{script_vs_agent}: a standard dialogue agent that accesses only its own persona profile and private goals, with no ToM analysis, no ensemble mechanism, and no multimodal perception---representing current LLM capabilities under realistic information asymmetry.
\textit{Script mode}~\cite{script_vs_agent}: a dialogue agent that receives \emph{both} agents' complete persona profiles and private goals simultaneously, serving as a full-information reference that violates realistic information constraints.

\noindent\textbf{Video Baselines.}
Comparisons are made with state-of-the-art talking head generation methods: Sonic~\cite{ji2025sonic}, an audio-driven approach without emotion control; and EDTalk~\cite{tan2025edtalk}, which supports discrete emotion categories with fixed pose constraints.

\noindent\textbf{Evaluation Protocol.}
\textit{Dialogue quality} is assessed using established LLM evaluation frameworks: Sotopia-Eval~\cite{sotopia} measures goal completion (believability, goal achievement, secret preservation, social rule adherence); LLM-Eval~\cite{llm-eval} assesses content quality, grammar, relevance, and appropriateness; GPT-Score~\cite{gpt-score} evaluates naturalness sub-dimensions including fluency, consistency (contradiction penalty against persona), coherence (topical continuity), depth (semantic richness of responses), diversity (lexical and strategic variety), and likeability; G-Eval~\cite{g-eval} provides complementary relevance, fluency, and coherence ratings. \textit{Audio quality} is measured by Open-D and Emo-Acc from URO-Bench~\cite{yan2025uro}. \textit{Visual synchronization quality} is measured by LipLMD, AVOffset, and AVConf following the ViCo protocol~\cite{zhou2022responsive}. \textit{Emotional expressiveness} is measured by Emo Score using an external emotion recognizer~\cite{RYUMINA2022435}. Sample evaluation prompts are provided in Supplementary~\ref{sec:implementation_details}.

\begin{table*}[!t]
    \centering
    \caption{\textbf{Dialogue evaluation results on the Persona-Scenario dataset} across goal completion and naturalness dimensions. Agent and Ours both operate under realistic information asymmetry where each agent only accesses its own persona and private goals. Script$^\dagger$ has full access to both sides' profiles and goals---an unrealistic condition absent in real interaction---and is shown in {\color{gray}gray} as a reference. \textbf{Bold} highlights the better result between the two realistic methods (Agent vs.\ Ours). Results are reported as mean $\pm$ std over 3 evaluation runs.}
    \label{tab:evaluation_metrics_mean_var}
    \vspace{-1ex}
    \footnotesize
    \setlength{\tabcolsep}{3pt}
    \renewcommand{\arraystretch}{0.88}
    \resizebox{\textwidth}{!}{%
    \begin{tabular}{lll rr rr rr}
    \toprule  
    \multirow{2}{*}{Dimension} & \multirow{2}{*}{Evaluator} & \multirow{2}{*}{Criteria} & \multicolumn{2}{c}{\color{gray}Script$^\dagger$} & \multicolumn{2}{c}{Agent} & \multicolumn{2}{c}{Ours} \\
    \cmidrule(lr){4-5} \cmidrule(lr){6-7} \cmidrule(lr){8-9}
     & & & Mean & Std & Mean & Std & Mean & Std \\
    \midrule  
    \multirow{4}{*}{Goal Completion} & \multirow{4}{*}{Sotopia-Eval~\cite{sotopia} $\uparrow$} &  
    Believability [0,10] & \color{gray}8.95 & \color{gray}0.11 & 8.17 & 0.13 & \textbf{9.19} & 0.15 \\
     & & Goal [0,10] & \color{gray}8.87 & \color{gray}0.06 & 6.54 & 0.03 & \textbf{7.92} & 0.05 \\
     & & Secret [-10,0] & \color{gray}-2.33 & \color{gray}0.04 & -1.77 & 0.08 & \textbf{-1.01} & 0.05 \\
     & & Social rules [-10,0] & \color{gray}-0.09 & \color{gray}0.04 & -0.07 & 0.03 & \textbf{-0.04} & 0.04 \\
     \midrule  
     & \multirow{4}{*}{LLM-Eval~\cite{llm-eval} $\uparrow$} & Content [0,100] & \color{gray}89.49 & \color{gray}0.57 & 82.98 & 0.53 & \textbf{88.23} & 0.89\\
     & & Grammar [0,100] & \color{gray}97.65 & \color{gray}0.14 & \textbf{98.11} & 0.22 & 97.13 & 0.27 \\
     & & Relevance [0,100] & \color{gray}94.72 & \color{gray}0.40 & 89.11 & 0.41 & \textbf{94.25} & 0.48 \\
     & & Appropriateness [0,100] & \color{gray}93.26 & \color{gray}0.45 & 88.29 & 0.26 & \textbf{94.98} & 0.61 \\
    \cmidrule{2-9}
    \multirow{5}{*}{Naturalness} & \multirow{6}{*}{GPT-Score~\cite{gpt-score} $\uparrow$} & Fluency [0,100] & \color{gray}98.55 & \color{gray}0.44 & 87.26 & 0.39 & \textbf{96.71} & 0.65 \\
     & & Consistency [0,100] & \color{gray}96.91 & \color{gray}0.29 & 85.23 & 0.32 & \textbf{97.51} & 0.49 \\
     & & Coherence [0,100] & \color{gray}98.29 & \color{gray}0.37 & 97.49 & 0.10 & \textbf{98.53} & 0.28 \\
     & & Depth [0,100] & \color{gray}55.85 & \color{gray}0.79 & 51.40 & 0.27 & \textbf{57.30} & 0.47 \\
     & & Diversity [0,100] & \color{gray}93.88 & \color{gray}0.42 & 73.20 & 0.76 & \textbf{96.33} & 0.50 \\
     & & Likeability [0,100] & \color{gray}91.02 & \color{gray}1.23 & 66.93 & 1.30 & \textbf{90.72} & 1.25 \\
    \cmidrule{2-9} 
     & \multirow{3}{*}{G-Eval~\cite{g-eval} $\uparrow$} & Relevance [1,5] & \color{gray}3.98 & \color{gray}0.06 & 3.52 & 0.05 & \textbf{3.95} & 0.07 \\
     & & Fluency [1,3] & \color{gray}2.52 & \color{gray}0.04 & 2.38 & 0.03 & \textbf{2.48} & 0.04 \\
     & & Coherent [1,5] & \color{gray}4.89 & \color{gray}0.06 & 4.79 & 0.05 & \textbf{4.85} & 0.05 \\\midrule  
    \multicolumn{9}{l}{\footnotesize $^\dagger$\,Full-information reference (unrealistic); does not participate in bold ranking.}
    \end{tabular}}
    \vspace{-2ex}
\end{table*}

\vspace{-2pt}
\subsection{LLM Social Simulation Results}

Table~\ref{tab:evaluation_metrics_mean_var} compares our system against an Agent mode baseline and a full-information Script mode across the goals completion and naturalness dimensions. Each cell reports the mean over three independent judge runs on the same set of generated dialogues, so the std captures evaluator variability rather than generation noise.

\noindent\textbf{Goal Completion.}
Our framework outperforms the Agent baseline in all Sotopia-Eval metrics. In believability, our method surpasses both Agent and Script modes, demonstrating that the ToM module and the Ensemble Mechanism (Sec.~\ref{sec:ensemble}) produce more authentic conversational behaviors. For goal completion, our method clearly outperforms the Agent baseline while operating under realistic information constraints. We note that Script mode outperforms our method on goal completion and fluency, which is expected given its unrestricted access to all private information. However, this unrestricted access comes at a cost: Script mode's poor secret preservation reveals that full information access leads to inadvertent leakage, while our Coherence Evaluator successfully enforces discretion under realistic constraints.

\noindent\textbf{Naturalness.}
Our framework achieves consistent improvements in naturalness metrics. The clearest gains appear in diversity and consistency, while depth places our method ahead of the Agent baseline and on par with the full-information Script mode, indicating that forced mental state inference yields responses at least as thoughtful as those produced under unrestricted information access. In terms of appropriateness and likeability, our method consistently outperforms the agent baseline, validating that the Social Reasoning Module guides responses toward appropriate dialogue. Consistency improvements confirm that the ensemble mechanism enforces persona stability across conversational turns.

\vspace{-2pt}
\subsection{Talking Head Generation Results}

\begin{figure*}[!t]
    \centering
    \includegraphics[width=\textwidth]{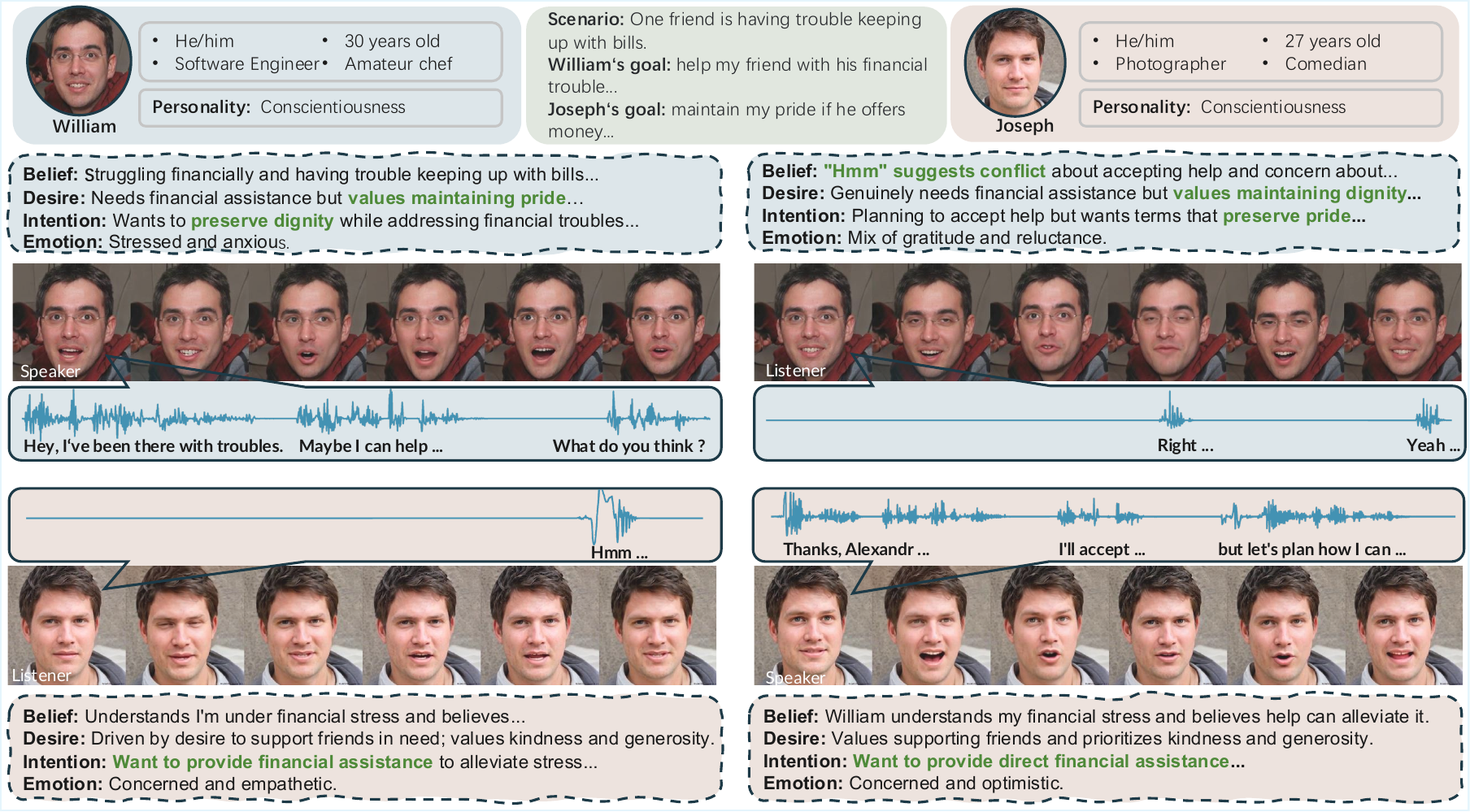}
    \vspace{-2ex}
    \caption{\textbf{Representative generated conversational avatars.} Each pair shows a speaker (left) and a reactive listener (right). The Perception module analyzes the partner's speech, emotion, and facial cues from the prior turn. The Social Reasoning module infers the partner's mental states via ToM and selects an appropriate response through the ensemble mechanism. The Expression module then renders the speaker's emotion-controlled speech and facial animation, alongside the listener's personality-driven backchannel reactions.}
    \label{fig:output_results}
    \vspace{-4ex}
\end{figure*}

We evaluate talking-head generation along three axes: audio quality, visual synchronization quality, and emotional expressiveness. Representative result are shown in Figure~\ref{fig:output_results}.

\begin{figure*}[!t]
    \centering
    \includegraphics[width=\textwidth]{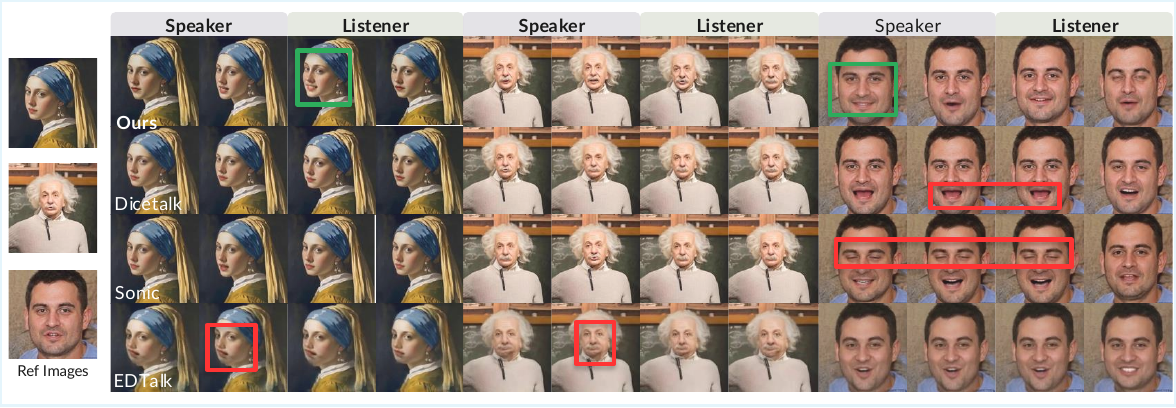}
    \vspace{-2ex}
    \caption{\textbf{Qualitative comparison with baseline methods.} \textcolor{red}{Red boxes} mark failure regions in baseline methods, including rigid/frozen facial behaviors (e.g., sustained closed eyes or persistently over-opened mouth) and lip distortion under occlusion. \textcolor{green!50!black}{Green boxes} highlight favorable regions from our method, showing responsive listener reactions and richer, more natural speaker emotional variation, including subtle smile states between neutral and happy. More video qualitative results are provided in Supplementary~\ref{sec:qualitative_results}.}
    \label{fig:qualitative_comparison}
    \vspace{-5ex}
\end{figure*}

\noindent\textbf{Comparison with SOTA Methods.}
Table~\ref{tab:sota_comparison} and Figure~\ref{fig:qualitative_comparison} summarize quantitative and qualitative comparisons with SOTA methods. Our method achieves the strongest overall trade-off on emotion accuracy and open-domain conversation capability. On the audio side, our method obtains the highest Open-D (61.00) and Emo-Acc (91.92), demonstrating that the dual-TTS strategy with emotion control produces both natural and emotionally accurate speech. We additionally include Ours (w/o listener) in Table~\ref{tab:sota_comparison}: removing listener generation degrades key metrics, supporting that listener modeling is a necessary component rather than an auxiliary add-on. EDTalk and Hallo3 achieve the best geometry-based scores due to their constrained pose and motion, which minimize landmark variance at the expense of expression dynamics. Our method prioritizes expression richness over rigid reconstruction and shows stronger overall performance than all other baselines. Since our control framework and dual-TTS strategy are independent of the underlying video generation model, our approach can be integrated with other video generation backbones.

\begin{table}[!t]
  \centering
  \footnotesize 
  \renewcommand{\arraystretch}{0.85} 
  \caption{\textbf{Audio and video quality comparison.} \textit{Video (Talking)} metrics evaluate the speaker only; \textit{Video (Full)} evaluates emotion over the full dual-agent video. EDTalk's fixed pose constraints reduce landmark variance, yielding better scores at the cost of expression dynamics; our method prioritizes expressiveness over rigid reconstruction. \textbf{Bold}: best per column.}
  \label{tab:sota_comparison}
  \vspace{-2mm} 
  \setlength{\tabcolsep}{2pt}
  \resizebox{\columnwidth}{!}{%
  \begin{tabular}{lcc ccc c}
  \toprule
  \multirow{2}{*}{Method} & \multicolumn{2}{c}{Audio} & \multicolumn{3}{c}{Video (Talking)} & Video (Full) \\
  \cmidrule(lr){2-3} \cmidrule(lr){4-6} \cmidrule(lr){7-7}
  & Open-D$\uparrow$ & Emo-Acc$\uparrow$ & LipLMD$\downarrow$ & AVOffset$\rightarrow$0 & AVConf$\uparrow$ & Emo Score$\uparrow$ \\
  \midrule
  SadTalker~\cite{SadTalker} & 48.50 & 58.00 & 26.50 & -37.18 & 2.45 & 0.28 \\
  Hallo3~\cite{Hallo3} & 55.00 & 65.00 & 32.00 & \textbf{-31.44} & 2.76 & 0.30 \\
  EDTalk~\cite{tan2025edtalk} & 57.83 & 75.11 & \textbf{21.91} & -31.54 & 2.49 & - \\
  Sonic~\cite{ji2025sonic} & 53.08 & 65.56 & 28.01 & -33.94 & 2.66 & - \\
  DICE-Talk~\cite{dice-talk} & 54.11 & 57.25 & 28.47 & -44.47 & \textbf{2.85} & 0.3057 \\
  Ours (w/o listener) & 57.50 & 90.28 & 33.72 & -38.67 & 2.64 & 0.3333 \\
  \textbf{Ours} & \textbf{61.00} & \textbf{91.92} & 22.39 & -33.96 & 2.50 & \textbf{0.3604} \\
  \bottomrule
  \end{tabular}}
  \vspace{-3mm} 
\end{table}

\vspace{-2pt}
\subsection{Ablation Study}
Table~\ref{tab:llm_ablation} reveals a clear modular separation across the two generation pathways. Removing the ensemble mechanism substantially drops Emo Score (0.36$\to$0.30) while leaving Emo-Acc nearly intact (91.92$\to$91.36), indicating that ensemble selection primarily drives video-level emotion control. Conversely, removing emotion control or perception sharply degrades Emo-Acc with only minor impact on Emo Score, confirming that these modules regulate audio-level expressiveness. Removing ToM causes broad degradation across all three metrics, reflecting its central role in coherent cross-modal generation and dialogue quality.

\begin{figure*}[!t]
\centering
\includegraphics[width=\textwidth]{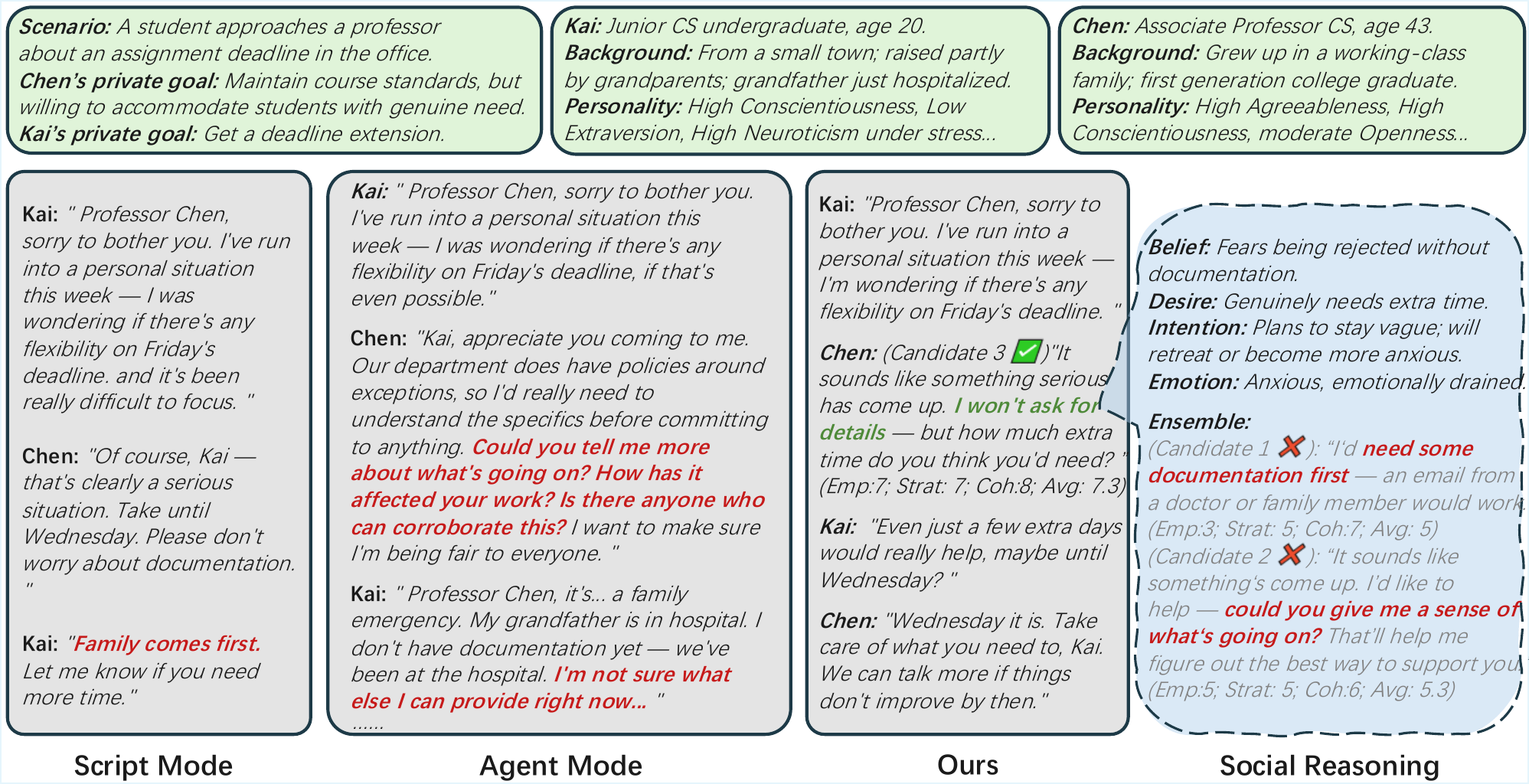}
\caption{\textbf{ToM causal case study.}
\textit{Script mode} violates information asymmetry: Agent \textit{Kai} only mentions ``a personal situation'' yet Agent \textit{Chen} responds ``Family comes first'' (shown in {\color{red}red}), referencing family information \textit{Kai} never disclosed.
\textit{Agent mode} produces verbose probing that cannot converge (shown in {\color{red}red}), forcing \textit{Kai} to over-disclose.
\textit{Ours:} the ToM module infers that \textit{Kai} fears being probed; the Ensemble mechanism rejects candidates that demand documentation or probe further (shown in {\color{red}red}) and selects the response that respects \textit{Kai}'s boundary (shown in {\color{green!60!black}green}) while still advancing \textit{Chen}'s goal.}
\label{fig:case_study}
\end{figure*}

\noindent\textbf{ToM Causal Analysis.}
Beyond aggregate metrics, we examine qualitative dialogue behavior in Figure~\ref{fig:case_study}.
Script mode frequently leaks private information that the partner never disclosed, producing socially implausible exchanges despite high goal-completion scores.
Agent mode tends to generate repetitive, verbose turns that fail to converge, as the lack of partner modeling leads to over-probing and formulaic responses.
Our method leverages multimodal perception and ToM inference to identify the partner's communicative boundary, then uses the ensemble mechanism to select a response strategy that respects that boundary, achieving goal completion while maintaining social plausibility under realistic information asymmetry. Additional qualitative results for dialogue generation and ToM analysis are provided in Supplementary~\ref{sec:qualitative_results}.

\vspace{-2pt}
\subsection{Generalization}
\label{subsec:generalization}

\noindent\textbf{Out-of-distribution scenarios.} To test whether the framework transfers beyond the Persona-Scenario dataset, we sample 30 cases from DialToM~\cite{DialToM} and Sotopia-Hard~\cite{script_vs_agent}, both curated for challenging social reasoning. As shown in Table~\ref{tab:ood}, performance degrades only modestly out of distribution, and relevance even improves (3.92 to 4.12). The main declines are in believability (9.25 to 7.35) and coherence (4.73 to 3.51). Goal achievement, secret preservation, and fluency stay close to their in-distribution values, indicating that ToM-driven reasoning rather than scenario memorization drives performance.

\begin{table}[t]
    \centering
    \footnotesize
    \renewcommand{\arraystretch}{0.9}
    \caption{\textbf{Generalization to out-of-distribution scenarios.} Metrics follow the Sotopia-Eval and G-Eval protocols: Goal, believability (Bel.), secret, and social-rules (Soc.) from Sotopia-Eval~\cite{sotopia}; coherence (Coh.), fluency (Flu.), and relevance (Rel.) from G-Eval~\cite{g-eval}. $\Delta$ denotes OOD minus in-distribution.}
    \label{tab:ood}
    \vspace{-1ex}
    \setlength{\tabcolsep}{8pt}
    \begin{tabular}{lccccccc}
    \toprule
    Setting & Goal$\uparrow$ & Bel.$\uparrow$ & Secret$\uparrow$ & Soc.$\uparrow$ & Coh.$\uparrow$ & Flu.$\uparrow$ & Rel.$\uparrow$ \\
     & {\scriptsize$[0,10]$} & {\scriptsize$[0,10]$} & {\scriptsize$[-10,0]$} & {\scriptsize$[-10,0]$} & {\scriptsize$[1,5]$} & {\scriptsize$[1,3]$} & {\scriptsize$[1,5]$} \\
    \midrule
    In-dist. & 8.00 & 9.25 & -1.00 & -0.05 & 4.73 & 2.45 & 3.92 \\
    OOD & 7.54 & 7.35 & -1.42 & -0.13 & 3.51 & 2.28 & 4.12 \\
    $\Delta$ & -0.46 & -1.90 & -0.42 & -0.08 & -1.22 & -0.17 & +0.20 \\
    \bottomrule
    \end{tabular}
    \vspace{-2ex}
\end{table}

\vspace{-2pt}
\subsection{User Study}
\label{subsec:user_study}

\noindent\textbf{Dialogue Quality.} Beyond the automatic metrics above, 22 raters scored dialogues sampled across the bargaining, movie selection, and dormitory conflict from Persona-Scenario dataset. Each dialogue was rated on five 1--5 scales that mirror the automatic dimensions, namely believability, goal achievement, secret preservation, depth, and fluency, plus an overall preference vote. As shown in Table~\ref{tab:dialog_human_eval}, our method outperforms the Agent baseline on four of the five scales and on overall preference. Qualitative dialogue comparisons are provided in Supplementary~\ref{sec:qualitative_results}.

\begin{table}[t]
    \centering
    \footnotesize
    \renewcommand{\arraystretch}{0.9}
    \caption{\textbf{Human evaluation of dialogue quality on the Persona-Scenario dataset.} \textbf{Bold}: better of the two realistic methods, Agent and Ours.}
    \label{tab:dialog_human_eval}
    \vspace{-1ex}
    \setlength{\tabcolsep}{10pt}
    \begin{tabular}{lcccccc}
    \toprule
    Method & Bel.$\uparrow$ & Goal$\uparrow$ & Secret$\uparrow$ & Depth$\uparrow$ & Flu.$\uparrow$ & Pref.$\uparrow$ \\
    \midrule
    {\color{gray}Script} & {\color{gray}3.70} & {\color{gray}3.82} & {\color{gray}3.49} & {\color{gray}3.54} & {\color{gray}3.95} & {\color{gray}40.5\%} \\
    Agent & 3.33 & 3.76 & 3.41 & \textbf{3.75} & 3.54 & 23.5\% \\
    Ours & \textbf{3.50} & \textbf{3.88} & \textbf{3.84} & 3.67 & \textbf{3.88} & \textbf{36.0\%} \\
    \bottomrule
    \end{tabular}
    \vspace{-2ex}
\end{table}

\noindent\textbf{Video Quality.} We conducted a user study with 85 participants. The 12 evaluation videos were sampled to cover 6 scenario types and 8 emotion categories, ensuring diverse representation. Participants rated each video on three 1--5 continuous scales: \textit{Emotional Expression} (1=no discernible emotion, 5=highly expressive), \textit{Communication Naturalness} (1=unnatural, 5=natural), and \textit{Overall Quality} (1=poor, 5=excellent). As shown in Table~\ref{tab:user_study}, our method achieves the highest scores across all three dimensions. The largest margin appears in emotional expression (4.50 vs.\ 3.38 for DICE-Talk), confirming that our emotion control pipeline produces perceptibly richer expressions. Our method also leads in naturalness, suggesting that dual-agent listener modeling contributes to perceived conversational realism. Notably, DICE-Talk scores competitively on naturalness (4.15) as the base video generator, yet lags on emotion (3.18), highlighting that visual fluency alone is insufficient without explicit emotion control.

\begin{table*}[!t]
  \begin{minipage}[t]{0.54\textwidth}
    \centering
    \caption{\textbf{Ablation study on Persona-Scenario dataset.}}
    \label{tab:llm_ablation}
    \vspace{-1mm}
    \footnotesize
    \resizebox{\linewidth}{!}{%
    \begin{tabular}{lccc}
    \toprule
    Ablation & Emo Score$\uparrow$ & Emo-Acc$\uparrow$ & Open-D$\uparrow$ \\
    \midrule
    (a) w/o ensemble & 0.3005 & 91.36$\pm$11.15 & 59.75$\pm$3.16 \\
    (b) w/o emotion & 0.3506 & 66.39$\pm$12.04 & 58.50$\pm$3.06 \\
    (c) w/o perception & 0.3547 & 68.81$\pm$13.11 & 58.08$\pm$2.63 \\
    (d) w/o ToM & 0.3003 & 65.84$\pm$13.71 & 58.25$\pm$2.91 \\
    \textbf{Ours} & \textbf{0.3604} & \textbf{91.92$\pm$9.93} & \textbf{61.00$\pm$3.01} \\
    \bottomrule
    \end{tabular}}
  \end{minipage}\hfill
  \begin{minipage}[t]{0.44\textwidth}
    \centering
    \caption{\textbf{User study.}}
    \label{tab:user_study}
    \vspace{-1mm}
    \footnotesize
    \resizebox{\linewidth}{!}{%
    \begin{tabular}{lccc}
    \toprule
    \multirow{2}{*}{Method} & Emotion$\uparrow$ & {Natural.$\uparrow$} & Overall.$\uparrow$ \\
     & Expr.[1-5] & [1-5] & Quality.[1-5] \\
    \midrule
    EDTalk~\cite{tan2025edtalk} & 3.38 & 2.13 & 2.08 \\
    Sonic~\cite{ji2025sonic} & 2.62 & 3.10 & 3.10 \\
    DICE-Talk~\cite{dice-talk} & 3.18 & 4.15 & 3.29 \\
    \textbf{Ours} & \textbf{4.50} & \textbf{4.34} & \textbf{3.38} \\
    \bottomrule
    \end{tabular}}
  \end{minipage}
  \vspace{-4mm}
\end{table*}

\vspace{-3pt}
\section{Conclusion}
\label{sec:conclusion}

We presented a closed-loop framework that bridges cognitive social reasoning and multimodal generation for conversational avatars. By integrating perception, Theory of Mind based social reasoning with an ensemble mechanism, and emotion controllable dual agent video synthesis, our system enables personality driven agents to perceive, reason about, and expressively respond to each other under realistic information asymmetry. Experiments show competitive or superior results on both dialogue quality and video generation. These findings suggest that equipping agents with explicit mental state inference can be more effective than simply granting unrestricted information access, pointing toward a promising direction for socially intelligent embodied agents.

\bibliographystyle{splncs04}
\bibliography{main}

\clearpage
\begin{center}
  {\Large\bfseries Resonant Minds: Closed-Loop Social Avatars with Theory of Mind}\\[0.6em]
  {\large\bfseries Supplementary Material}
\end{center}
\vspace{1.2em}
\noindent In this supplementary material, we present additional implementation details in Sec.~\ref{sec:implementation_details}, comprehensive persona-scenario dataset creation details in Sec.~\ref{sec:dataset_creation}, experiment details with extended results in Sec.~\ref{sec:experiment_details}, and qualitative results with in-depth discussion in Sec.~\ref{sec:qualitative_results}.

\renewcommand{\thesection}{\Alph{section}}
\setcounter{section}{0}
\renewcommand{\thesubsection}{\thesection.\arabic{subsection}}
\renewcommand{\thesubsubsection}{\thesubsection.\arabic{subsubsection}}
\renewcommand{\thefigure}{\thesection.\arabic{figure}}
\renewcommand{\thetable}{\thesection.\arabic{table}}
\renewcommand{\theequation}{\thesection.\arabic{equation}}

\renewcommand{\theHsection}{supp.\thesection}
\renewcommand{\theHsubsection}{supp.\thesubsection}
\renewcommand{\theHsubsubsection}{supp.\thesubsubsection}
\renewcommand{\theHfigure}{supp.\thefigure}
\renewcommand{\theHtable}{supp.\thetable}
\renewcommand{\theHequation}{supp.\theequation}

\section{Implementation Details}
\label{sec:implementation_details}
\setcounter{figure}{0}
\setcounter{table}{0}
\setcounter{equation}{0}

This section provides comprehensive implementation details. We describe the model architecture in Sec.~\ref{subsec:model_architecture}, prompt templates in Sec.~\ref{subsec:prompt_templates}, and evaluation metrics in Sec.~\ref{subsec:evaluation_metrics}. All experiments are conducted on servers equipped with 8 NVIDIA V100 GPUs, using PyTorch 2.6.0 and CUDA 12.2.

\subsection{Model Architecture}
\label{subsec:model_architecture}

\subsubsection{Perception Module}

\noindent We adopt HumanOmni-7B~\cite{humanomni} as the unified multimodal perception model, achieving joint understanding of video, audio, and text at 7B parameter scale. During inference, we configure the model with \texttt{do\_sample\allowbreak=False} for deterministic outputs, \texttt{temperature}$\,{=}\,$\texttt{1.0}, and \texttt{max\_new\_tokens}$\,{=}\,$\texttt{512}. Video inputs are processed at 1 FPS sampling rate to capture key frames while maintaining efficiency. The perception module runs on a dedicated GPU to enable parallel processing with other pipeline components.

Perception analysis comprises three sub-tasks---Emotion Analysis, Facial Expression Analysis, and Speech Recognition---all adopting first-person observational perspective to reinforce the subjective experience of agent mode. Detailed prompt templates are provided in Sec.~\ref{subsec:prompt_templates}.

\subsubsection{Social Reasoning Module}

\noindent We employ GPT-4o~\cite{achiam2023gpt} as the core model for dialogue generation and reasoning via the OpenAI API. For dialogue generation, we set \texttt{temperature}$\,{=}\,$\texttt{0.7} with \texttt{max\_tokens}$\,{=}\,$\texttt{500}. For structured reasoning tasks, we use \texttt{temperature}$\,{=}\,$\texttt{0.3} and enable JSON mode to enforce structured outputs. All reasoning modules employ chain-of-thought prompting to elicit step-by-step deliberation.

\begin{description}[leftmargin=1.5em,labelindent=1.5em,labelsep=0.4em,itemsep=3pt,topsep=3pt,parsep=0pt,font=\normalfont\bfseries]
\item[Theory of Mind Analysis]
The ToM module uses GPT-4o for mental state inference (\texttt{temperature}$\,{=}\,$\texttt{0.3}, \texttt{max\_tokens}$\,{=}\,$\texttt{800}). It constructs prompts based on the BDIE framework to infer the partner's hidden mental states from observable behaviors (see Sec.~\ref{subsec:prompt_templates} for the detailed template). Each BDIE dimension is limited to 150 words. ToM analysis begins from the second turn onwards, as the first turn lacks sufficient observation history. Results are returned in JSON format with fields: \textit{belief}, \textit{desire}, \textit{intention}, and \textit{emotion}.

\item[Candidate Generator]
The generator produces $K{=}3$ diverse candidate responses using \texttt{temperature}$\,{=}\,$\texttt{0.9} and \texttt{max\_tokens}$\,{=}\,$\texttt{500}, encouraging variation in communication styles and strategic approaches.

\item[Ensemble Mechanism]
Each evaluator (Empathy, Strategic, Coherence; see Sec.~3.3 of the main paper) uses GPT-4o with \texttt{temperature\allowbreak=0.3} and \texttt{max\_new\allowbreak\_tokens\allowbreak=200}. Scoring criteria: 8--10 for exceptional, 5--7 for adequate, 0--4 for weak. Equal weights $\lambda_{\text{emp}} = \lambda_{\text{strat}} = \lambda_{\text{coh}} = 1/3$ are used by default; sensitivity analysis is provided in Sec.~\ref{subsec:ensemble_sensitivity}.
\end{description}

\subsubsection{Expression Module}

\begin{description}[leftmargin=1.5em,labelindent=1.5em,labelsep=0.4em,itemsep=3pt,topsep=3pt,parsep=0pt,font=\normalfont\bfseries]
\item[Text-to-Emotion Adapter]
The adapter maps natural language emotion descriptions to 8-dimensional weight vectors $\mathbf{w} \in \mathbb{R}^8$ for controlling talking head generation. It consists of a frozen CLIP text encoder followed by a trainable MLP mapping network. The final softmax layer produces valid probability distributions ($\sum_{i} w_i = 1$, $w_i \geq 0$) over eight basic emotion categories: contempt, sadness, happiness, surprise, anger, disgust, fear, and neutral.

The adapter is trained on ${\sim}360$ synthetic text-weight pairs generated through template-based generation. Each sample is manually verified to ensure the target weight vector faithfully reflects the textual description. Training samples are carefully designed to cover three complementary categories: single emotions with varying intensities (e.g., ``mild sadness'' $\to$ ``overwhelming grief''), intensity modulations spanning mild to extreme ranges, and mixed emotions combining two affective states with explicitly distributed weight allocations (e.g., ``bittersweet happiness'' $\to$ $[0,0.3,0.5,0,0,0,0,0.2]$). This curated diversity ensures the adapter learns a smooth, well-calibrated mapping rather than memorizing a sparse set of prototypes. Only the MLP is trained while CLIP remains frozen, enabling generalization to unseen emotion expressions without large-scale annotated data. Architecture and training hyperparameters are summarized in Table~\ref{tab:adapter_config}.

\begin{table}[!htbp]
  \centering
  \caption{\textbf{Text-to-Emotion Adapter configuration.} The adapter maps free-form emotion descriptions to an 8-dimensional probability vector $\mathbf{w}\in\mathbb{R}^8$ over basic emotion categories. Only the MLP layers are trained; the CLIP text encoder remains frozen throughout.}
  \label{tab:adapter_config}
  \small
  \begin{tabular}{ll}
  \toprule
  \multicolumn{2}{l}{\textit{Architecture}} \\
  \midrule
  Text encoder & CLIP-ViT-Large-Patch14 (frozen) \\
  Embedding dimension & 768 \\
  MLP hidden layers & 2 (dim 256 each) \\
  Normalization & LayerNorm \\
  Activation & ReLU \\
  Dropout & 0.1 \\
  Output dimension & 8 (softmax-normalized) \\
  \midrule
  \multicolumn{2}{l}{\textit{Training}} \\
  \midrule
  Loss function & KL divergence \\
  Optimizer & AdamW \\
  Learning rate & $1 \times 10^{-3}$ \\
  LR schedule & Cosine annealing $\to 1 \times 10^{-6}$ \\
  Epochs & 30 \\
  Batch size & 16 \\
  Gradient clipping & Enabled \\
  Training samples & ${\sim}360$ synthetic pairs \\
  \bottomrule
  \end{tabular}
\end{table}

\item[Index-TTS v2]
Speech synthesis employs Index-TTS v2~\cite{zhou2025indextts2breakthroughemotionallyexpressive} with 8-dimensional emotion vector control and intensity coefficient $\alpha{=}1.5$.

\item[DICE-Talk Video Generation]
Video generation employs DICE-Talk~\cite{dice-talk}, fine-tuned on SVD-xt~\cite{svd} using FP16 precision. DICE-Talk natively accepts only a discrete selection from 8 emotion categories; our text-to-emotion adapter bridges this gap by converting free-form natural language descriptions into the required 8-dimensional weight vector, enabling continuous and fine-grained emotion control that the original model cannot achieve on its own. The full configuration is listed in Table~\ref{tab:dicetalk_config}.

\begin{table}[!htbp]
  \centering
  \caption{\textbf{DICE-Talk video generation configuration.}}
  \label{tab:dicetalk_config}
  \small
  \begin{tabular}{ll}
  \toprule
  Parameter & Value \\
  \midrule
  Minimum resolution & $512\times512$ \\
  Inference steps & 25 \\
  Appearance guidance scale (\texttt{ref\_scale}) & 3.0 \\
  Emotion guidance scale (\texttt{emo\_scale}) & 6.0 \\
  Frame rate & 12.5 FPS \\
  Frames per batch & 25 \\
  Motion bucket ID & 8 (standard), 16 (expression) \\
  Noise augmentation & 0.0 \\
  Precision & FP16 \\
  \bottomrule
  \end{tabular}
\end{table}
\end{description}

\subsection{Prompt Templates}
\label{subsec:prompt_templates}

We provide the key prompt templates used across our framework in Tables~\ref{tab:perception_prompts} and~\ref{tab:prompt_layers}.

\begin{table}[!htbp]
\centering
\caption{Perception and Theory of Mind prompt templates. Placeholders in \texttt{braces} are filled at runtime.}
\label{tab:perception_prompts}
\renewcommand{\arraystretch}{1.35}
\setlength{\tabcolsep}{4pt}
\begin{tabular}{@{}l l p{7.2cm}@{}}
\toprule
\textbf{Module} & \textbf{Task} & \textbf{Prompt Template} \\
\midrule
\multirow{3}{*}{Perception}
  & Emotion   & \textit{``What's the major emotion of \texttt{\{name\}}?''} \\
  & Expression & \textit{``Describe \texttt{\{name\}} and the major action in detail.''} \\
  & Speech    & \textit{``What did \texttt{\{name\}} say?''} \\
\midrule
ToM & BDIE & \textit{``Based on \texttt{\{partner\}}'s speech, emotion, and facial expression, infer: (1)~Belief: what they believe about the situation; (2)~Desire: what they want to achieve; (3)~Intention: their planned next action; (4)~Emotion: underlying emotional state and reasons.''} \\
\bottomrule
\end{tabular}
\end{table}

\begin{table}[!htbp]
\centering
\caption{Three-layer prompt architecture for dialogue generation.}
\label{tab:prompt_layers}
\renewcommand{\arraystretch}{1.35}
\setlength{\tabcolsep}{4pt}
\begin{tabular}{@{}l p{9.5cm}@{}}
\toprule
\textbf{Layer} & \textbf{Prompt Template} \\
\midrule
System & \textit{``You are \texttt{\{agent\_name\}}, a \texttt{\{role\}}. You are having a face-to-face conversation with \texttt{\{partner\_name\}}. You can ONLY access your own profile and goals. Do NOT assume any knowledge of the other party's private information. Respond naturally in first person.''} \\
\midrule
Core & \textit{``[Persona Profile] \texttt{\{persona\}} \newline [Scenario] \texttt{\{scenario\_description\}} \newline [Your Private Goal] \texttt{\{private\_goal\}}''} \\
\midrule
Empirical & \textit{``[Dialogue History] \texttt{\{history\}} \newline [Perception] Speech: \texttt{\{speech\}}; Emotion: \texttt{\{emotion\}}; Expression: \texttt{\{expression\}} \newline [ToM Analysis] B: \texttt{\{belief\}}; D: \texttt{\{desire\}}; I: \texttt{\{intention\}}; E: \texttt{\{emo\}} \newline [Past Ensemble Decisions] \texttt{\{ensemble\_log\}}''} \\
\bottomrule
\end{tabular}
\end{table}

\section{Persona-Scenario Dataset Creation}
\label{sec:dataset_creation}
\setcounter{figure}{0}
\setcounter{table}{0}
\setcounter{equation}{0}

This section provides detailed descriptions of the Persona-Scenario dataset construction pipeline introduced in Sec.~3.5 of the main paper. The overall pipeline consists of four stages: (1)~constructing 50 psychologically-grounded persona profiles through a Sparse-to-Rich generation pipeline; (2)~adapting 90 social scenarios with private goals from Sotopia~\cite{sotopia}; (3)~curating 365 face-voice avatar pairs for multimodal rendering; and (4)~assembling reproducible test cases by pairing personas, scenarios, and avatars.

\subsection{Persona Profile Construction}

We construct 50 personas using a Sparse-to-Rich pipeline comprising two layers, followed by private goal assignment at test time.

\noindent\textbf{Facts Layer} Following the persona expansion approach from~\cite{faithful-persona}, each persona begins with a manually authored seed consisting of 3--5 key attributes (e.g., age, gender, occupation, location). GPT-4o then expands these seeds into a full set of biographical facts---including education, family status, hobbies, and life experiences. To ensure logical coherence, we employ a T5-based Natural Language Inference (NLI) model for consistency checking: each newly generated attribute is validated against existing facts, and contradictory statements are removed. The generation process iterates until the profile converges to a self-consistent set. The validated attributes are synthesized into a coherent biographical narrative. For example:
\begin{quote}
\small\textit{``34-year-old female architect based in Chicago, divorced, passionate about sustainable design. She holds a Master's degree from IIT and runs a small firm specializing in eco-friendly residential projects.''}
\end{quote}

\noindent\textbf{Psychological Layer} We apply the PERSONALITY PROMPTING method from~\cite{eval-big-five} to generate rich personality narratives grounded in the Big Five model~\cite{de2002big}. To ensure balanced coverage, we pre-assign each persona a target Big Five configuration by sampling trait levels (high/medium/low) such that the overall distribution across 50 personas is approximately uniform per dimension. For each target configuration, we extract psychology-informed descriptive keywords, then prompt GPT-4o to generate detailed personality portraits. For example, a persona with high Openness and low Neuroticism yields:
\begin{quote}
\small\textit{``She approaches conflicts with curiosity rather than defensiveness and readily explores unconventional solutions. Her emotional stability allows her to remain calm under pressure, making her a reliable mediator in tense negotiations.''}
\end{quote}

\noindent\textbf{Private Social Goals} Goals $\mathcal{G}_\mathfrak{i}$ are not fixed per persona but generated dynamically at test time, conditioned on both the persona profile and the assigned scenario. This ensures goals are contextually appropriate (e.g., a business-savvy persona in a negotiation scenario receives a profit-maximization goal with specific price constraints). Goals for the two interacting agents are generated independently to maintain information asymmetry---neither agent has access to the partner's objective.

Final persona profiles are stored in JSON format with three core components: $P_\mathfrak{i} = (\mathcal{B}_\mathfrak{i}, \Psi_\mathfrak{i}, \mathcal{G}_\mathfrak{i})$, representing background, personality traits, and private social goals respectively.

\subsection{Scenario Construction}

We adapt 90 social scenarios from the Sotopia~\cite{sotopia} dataset, which originally contains over 150 scenarios. We select 90 that satisfy two criteria: (1)~suitability for two-party interaction (excluding group scenarios), and (2)~coverage of all six primary interaction types---Negotiation, Exchange, Competition, Collaboration, Accommodation, and Persuasion---with 15 scenarios per type. Each scenario comprises:
\begin{itemize}[nosep,leftmargin=1.5em]
  \item \textbf{Scenario description}: the interaction setting, background context, and environmental constraints.
  \item \textbf{Private social goals}: each party receives an independent objective with specific success criteria (e.g., ``Sell the laptop for at least \$500'' vs.\ ``Buy the laptop for under \$400''), ensuring information asymmetry.
\end{itemize}

\noindent Scenario-persona pairing is accomplished through semantic matching: we use GPT-4o to score compatibility between persona backgrounds and scenario settings, selecting pairs where the persona's occupation, personality, and life experience are plausible for the given context. For example, a scenario set in a real estate office is preferentially paired with personas who have business or finance backgrounds.

\subsection{Face-Voice Avatar Pairs}

We curate 365 face-voice pairs (230 male, 135 female) from the VICO~\cite{zhou2022responsive} and VICOX~\cite{vicox} public datasets. Each pair contains a $512\times512$ high-resolution portrait image (used as DICE-Talk reference) and a 3--5 second reference audio clip (used as Index-TTS speaker prompt). All face images undergo preprocessing including face detection, alignment, and center-cropping. Gender labels are manually annotated for persona-avatar gender matching.

\subsection{Test Case Generation}

For reproducibility, all experiments use 10 identical test cases generated with seed 2132. Each test case is assembled through: (1)~Randomly sample two personas from the 50-persona pool, ensuring compatible backgrounds; (2)~Select a scenario via the semantic matching described above; (3)~Assign gender-aligned face-voice pairs from the 365-pair pool; (4)~Generate private social goals for each persona tailored to the selected scenario, using GPT-4o with explicit constraints on information asymmetry and goal conflict/cooperation dynamics. This controlled test set enables fair comparison across all baseline methods and ablation variants.

\section{Experiment Details and Extended Results}
\label{sec:experiment_details}
\setcounter{figure}{0}
\setcounter{table}{0}
\setcounter{equation}{0}

\subsection{Evaluation Metrics}
\label{subsec:evaluation_metrics}
Our evaluation framework assesses system performance across three modalities.
\subsubsection{Dialogue Quality Metrics}

We employ four complementary automated evaluation frameworks, all using GPT-5 as the judge. \textbf{Sotopia-Eval}~\cite{sotopia} is a multi-dimensional framework rooted in sociology and psychology that evaluates goal-driven social interactions beyond binary task completion. \textbf{LLM-Eval}~\cite{llm-eval} is a unified single-prompt schema that simultaneously outputs multi-dimensional quality scores via structured JSON output in a single model call. \textbf{GPT-Score}~\cite{gpt-score} is a training-free framework that leverages conditional generation probability to score text, using natural language instructions to customize evaluation dimensions on the fly. \textbf{G-Eval}~\cite{g-eval} uses Chain-of-Thought reasoning to auto-generate evaluation steps, then extracts token probabilities for weighted scoring, yielding continuous scores with higher human correlation than discrete integer scoring. All metrics are summarized in Table~\ref{tab:dialogue_metrics}.

\begin{table}[!htbp]
  \centering
  \caption{\textbf{Dialogue quality metrics.} Four frameworks covering social interaction quality and linguistic naturalness.}
  \label{tab:dialogue_metrics}
  \small
  \renewcommand{\arraystretch}{1.15}
  \setlength{\tabcolsep}{5pt}
  \begin{tabular}{@{}llcl@{}}
  \toprule
  \textbf{Framework} & \textbf{Metric} & \textbf{Range} & \textbf{Description} \\
  \midrule
  \multirow{4}{*}{\shortstack[l]{Sotopia-\\Eval~\cite{sotopia}}}
  & Believability       & [0, 10]     & Character consistency and realism \\
  & Goal Achieve.       & [0, 10]     & Private social goal completion \\
  & Secret Pres.        & [$-$10, 0]  & Avoidance of private info leakage \\
  & Social Rules        & [$-$10, 0]  & Adherence to social norms \\
  \midrule
  \multirow{4}{*}{\shortstack[l]{LLM-\\Eval~\cite{llm-eval}}}
  & Content             & [0, 100]    & Informativeness and substance \\
  & Grammar             & [0, 100]    & Grammatical correctness \\
  & Relevance           & [0, 100]    & Topical relevance to context \\
  & Appropriateness     & [0, 100]    & Social and contextual suitability \\
  \midrule
  \multirow{6}{*}{\shortstack[l]{GPT-\\Score~\cite{gpt-score}}}
  & Fluency             & [0, 100]    & Linguistic smoothness \\
  & Consistency         & [0, 100]    & Persona contradiction penalty \\
  & Coherence           & [0, 100]    & Topical continuity across turns \\
  & Depth               & [0, 100]    & Semantic richness of responses \\
  & Diversity           & [0, 100]    & Lexical and strategic variety \\
  & Likeability         & [0, 100]    & Perceived friendliness and warmth \\
  \midrule
  \multirow{3}{*}{\shortstack[l]{G-\\Eval~\cite{g-eval}}}
  & Relevance           & [1, 5]      & Response--context alignment \\
  & Fluency             & [1, 3]      & Readability and naturalness \\
  & Coherence           & [1, 5]      & Logical structure and flow \\
  \bottomrule
  \end{tabular}
\end{table}

\subsubsection{Audio and Video Quality Metrics}

For \textbf{audio quality}, we adopt metrics from URO-Bench~\cite{yan2025uro}, a comprehensive S2S benchmark that evaluates spoken dialogue models across understanding, reasoning, and oral conversation dimensions. For \textbf{video quality}, we follow the ViCo Challenge evaluation protocol~\cite{zhou2022responsive}, which categorizes metrics into speaker-specified and listener-specified dimensions. We additionally employ Emo-Score using a pretrained facial emotion recognition model~\cite{RYUMINA2022435} to validate cross-modal emotion consistency. All metrics are summarized in Table~\ref{tab:av_metrics}.

\begin{table}[!htbp]
  \centering
  \caption{\textbf{Audio and video quality metrics.} Speaker metrics follow the ViCo Challenge protocol~\cite{zhou2022responsive}; audio metrics are from URO-Bench~\cite{yan2025uro}.}
  \label{tab:av_metrics}
  \small
  \renewcommand{\arraystretch}{1.15}
  \setlength{\tabcolsep}{6pt}
  \begin{tabular}{@{}llcl@{}}
  \toprule
  \textbf{Scope} & \textbf{Metric} & \textbf{Range} & \textbf{Description} \\
  \midrule
  \multirow{2}{*}{Audio}
  & Open-D $\uparrow$    & [0, 100]  & Open-domain response quality (GPT-4o judge) \\
  & Emo-Acc $\uparrow$   & [0, 100]  & Intended vs.\ perceived emotion alignment \\
  \midrule
  \multirow{3}{*}{Speaker}
  & LipLMD $\downarrow$       & --  & Mouth landmark distance (lip sync) \\
  & AVOffset $\rightarrow$0   & --  & SyncNet temporal offset (0 = perfect) \\
  & AVConf $\uparrow$         & --  & SyncNet audio-visual confidence \\
  \midrule
  \multirow{2}{*}{Listener}
  & ExpFD $\downarrow$   & --  & 3DMM expression Fr\'{e}chet distance \\
  & PoseFD $\downarrow$  & --  & 3DMM head pose Fr\'{e}chet distance \\
  \midrule
  Full video
  & Emo-Score $\uparrow$ & [0, 1]  & Emotion classifier confidence on target emotion \\
  \bottomrule
  \end{tabular}
\end{table}

\subsection{Baselines}
\label{subsec:baselines}

\subsubsection{Dialogue Generation Baselines}

We compare against two dialogue baselines operating under different information access paradigms. Table~\ref{tab:dialogue_baselines} summarizes the key differences. Both baselines use GPT-4o with identical generation parameters.

\noindent\textbf{Agent Mode}~\cite{script_vs_agent} represents a standard LLM dialogue agent under realistic information asymmetry---each agent accesses only its own persona and goals. Without ToM, ensemble, or multimodal perception, the agent relies solely on text-based dialogue history for response generation. This represents current LLM capabilities under realistic constraints.

\noindent\textbf{Script Mode}~\cite{script_vs_agent} provides a full-information reference where a single LLM generates dialogue for both participants with omniscient access to all private information. While this removes the inference challenge and enables rapid convergence, it paradoxically leads to worse secret preservation, as agents inadvertently leak private information when they know the partner's goals. This baseline quantifies the performance gap introduced by information asymmetry.

\begin{table}[!htbp]
  \centering
  \caption{\textbf{Dialogue generation baseline comparison.} Module availability and information access across the three settings. \emph{$K$}: number of candidate responses generated per turn before ensemble selection. \emph{Prompt layers}: depth of the prompt hierarchy---2 layers comprise persona profile and dialogue history; the 3rd layer in our system adds ToM-derived mental state context and multimodal perception results.}
  \label{tab:dialogue_baselines}
  \small
  \renewcommand{\arraystretch}{1.15}
  \setlength{\tabcolsep}{4pt}
  \begin{tabular}{@{}lccc@{}}
  \toprule
  \textbf{Property} & \textbf{Agent} & \textbf{Script} & \textbf{Ours} \\
  \midrule
  LLM backbone           & GPT-4o & GPT-4o & GPT-4o \\
  Temperature             & 0.7    & 0.7    & 0.7 \\
  Max tokens              & 500    & 500    & 500 \\
  \midrule
  Info.\ access           & Own only & Both agents & Own only \\
  Multimodal perception   & \texttimes & \texttimes & \checkmark \\
  ToM analysis            & \texttimes & \texttimes & \checkmark \\
  Candidate generation    & \texttimes & \texttimes & \checkmark ($K{=}3$) \\
  Ensemble mechanism      & \texttimes & \texttimes & \checkmark \\
  Prompt layers           & 2      & 2      & 3 \\
  \bottomrule
  \end{tabular}
\end{table}

\subsubsection{Talking Head Generation Baselines}

We compare against state-of-the-art talking head methods. For fair comparison, all methods use Index-TTS v2 for audio synthesis and are evaluated on identical test cases. Table~\ref{tab:video_baselines} summarizes the configurations.

\begin{table}[!htbp]
  \centering
  \caption{\textbf{Talking head baseline comparison.} Key capabilities across video generation methods. \emph{Emotion control}: discrete methods select from a fixed label set; continuous methods use an 8-dimensional weight vector over emotion categories. \emph{Reactive listener}: whether the method generates context-aware listener facial responses during the partner's speaking turn.}
  \label{tab:video_baselines}
  \small
  \renewcommand{\arraystretch}{1.15}
  \resizebox{\columnwidth}{!}{%
  \begin{tabular}{@{}lcccccc@{}}
  \toprule
  \textbf{Property} & \textbf{EDTalk} & \textbf{Sonic} & \textbf{SadTalker} & \textbf{HunyuanPort.} & \textbf{DICE-Talk} & \textbf{Ours} \\
  \midrule
  Resolution         & $512^2$     & $512^2$     & $256^2$      & $512^2$       & $512^2$     & $512^2$ \\
  Emotion control    & Discrete-8  & None        & None         & Audio-guided  & Continuous-8 & Continuous-8 \\
  Emotion source     & Keyword map & Audio prosody & Audio+3DMM & Audio+text    & Txt-to-Emo   & Txt-to-Emo \\
  Reactive listener  & \texttimes  & \texttimes  & \texttimes   & \texttimes    & \texttimes  & \checkmark \\
  Fixed head pose    & \checkmark  & \texttimes  & \texttimes   & \texttimes    & \texttimes  & \texttimes \\
  Social reasoning   & \texttimes  & \texttimes  & \texttimes   & \texttimes    & \texttimes  & \checkmark \\
  \bottomrule
  \end{tabular}%
  }
\end{table}

\noindent\textbf{EDTalk}~\cite{tan2025edtalk} is an autoencoder-based framework that decomposes facial dynamics into three disentangled latent spaces (mouth, pose, expression) via orthogonal learnable bases. In audio-driven mode, expression weights are predicted from HuBERT audio features and EmoBERTA text embeddings, supporting 8 discrete emotion categories from the MEAD dataset.

\noindent\textbf{Sonic}~\cite{ji2025sonic} is a diffusion-based portrait animation method built on SVD that emphasizes global audio perception rather than explicit emotion conditioning. Facial expressions are implicitly driven by speech prosody through context-enhanced audio learning, while a motion-decoupled controller separately governs head movement amplitude and expression strength via two motion-bucket parameters. Since Sonic provides no emotion input interface, expressions emerge solely from the audio signal, limiting fine-grained affective control. It does not support reactive listener generation.

\noindent\textbf{DICE-Talk}~\cite{dice-talk} is a diffusion-based emotional talking head method also built on SVD. It extracts disentangled emotion priors via a cross-modal audio-visual embedder and injects them into the diffusion UNet through a correlation-enhanced emotion bank. Critically, DICE-Talk natively accepts only one of 8 fixed emotion prior vectors---mean embeddings precomputed from MEAD training data for each discrete category (happy, sad, angry, surprised, disgusted, contempt, fear, neutral). It does not support free-form text or continuous emotion input. In our pipeline, the text-to-emotion adapter bridges this gap by mapping natural language descriptions to continuous 8-dimensional weight vectors over these categories.

\subsection{User Study Protocol}
\label{subsec:user_study_protocol}

We recruited 85 participants (42 male, 43 female, aged 22--45, mean age 28.3) from university mailing lists and online platforms, spanning computer science (35\%), humanities and social sciences (40\%), and other fields (25\%). The study was IRB-approved (\#2024-CV-089), and each participant received \$15 for a 30-minute session.

\begin{figure}[!htbp]
\centering
\includegraphics[width=0.82\columnwidth]{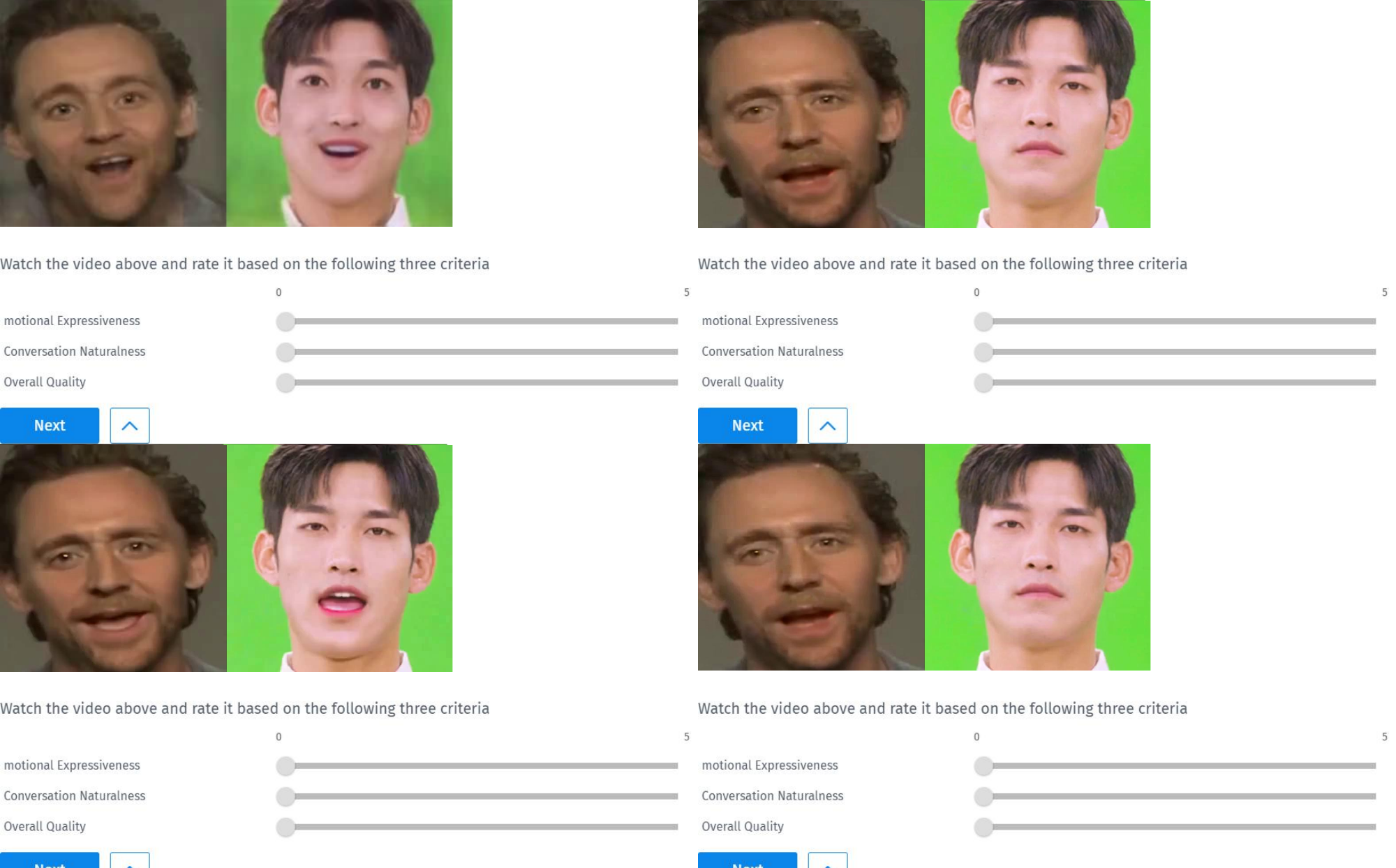}
\caption{\textbf{User study interface.} Participants view side-by-side video pairs in randomized positions and rate three dimensions via continuous sliders.}
\label{fig:user_study_interface}
\end{figure}

Using a within-subjects design, each participant evaluated all four methods (Ours, EDTalk, Sonic, DICE-Talk) in randomized order on 12 representative video clips of 20--30 seconds, covering 6 scenario types, 8 emotion categories, and diverse personality traits. As shown in Fig.~\ref{fig:user_study_interface}, videos were displayed side-by-side with randomized left-right positions, and participants rated Emotional Expressiveness, Conversation Naturalness, and Overall Quality on continuous 0--5 sliders. We embedded three repeated videos as attention checks and excluded participants with test-retest correlation below 0.7. After filtering, 82 valid participants remained.

\noindent\textbf{Statistical Analysis}
We performed repeated-measures analysis of variance on the collected ratings, which tests whether the score differences across methods are statistically significant while accounting for within-participant variability. All three dimensions showed significant main effects ($F>45.2$, $p<0.001$), and pairwise comparisons with Bonferroni correction confirmed that our method outperformed every baseline ($p<0.001$). Cohen's $d$ effect sizes were 1.82 against EDTalk, 2.14 against Sonic, and 1.23 against DICE-Talk, all well above the 0.8 threshold for a large effect.

\subsection{Ensemble Weight Sensitivity}
\label{subsec:ensemble_sensitivity}

\begin{figure}[!htbp]
\centering
\includegraphics[width=\columnwidth]{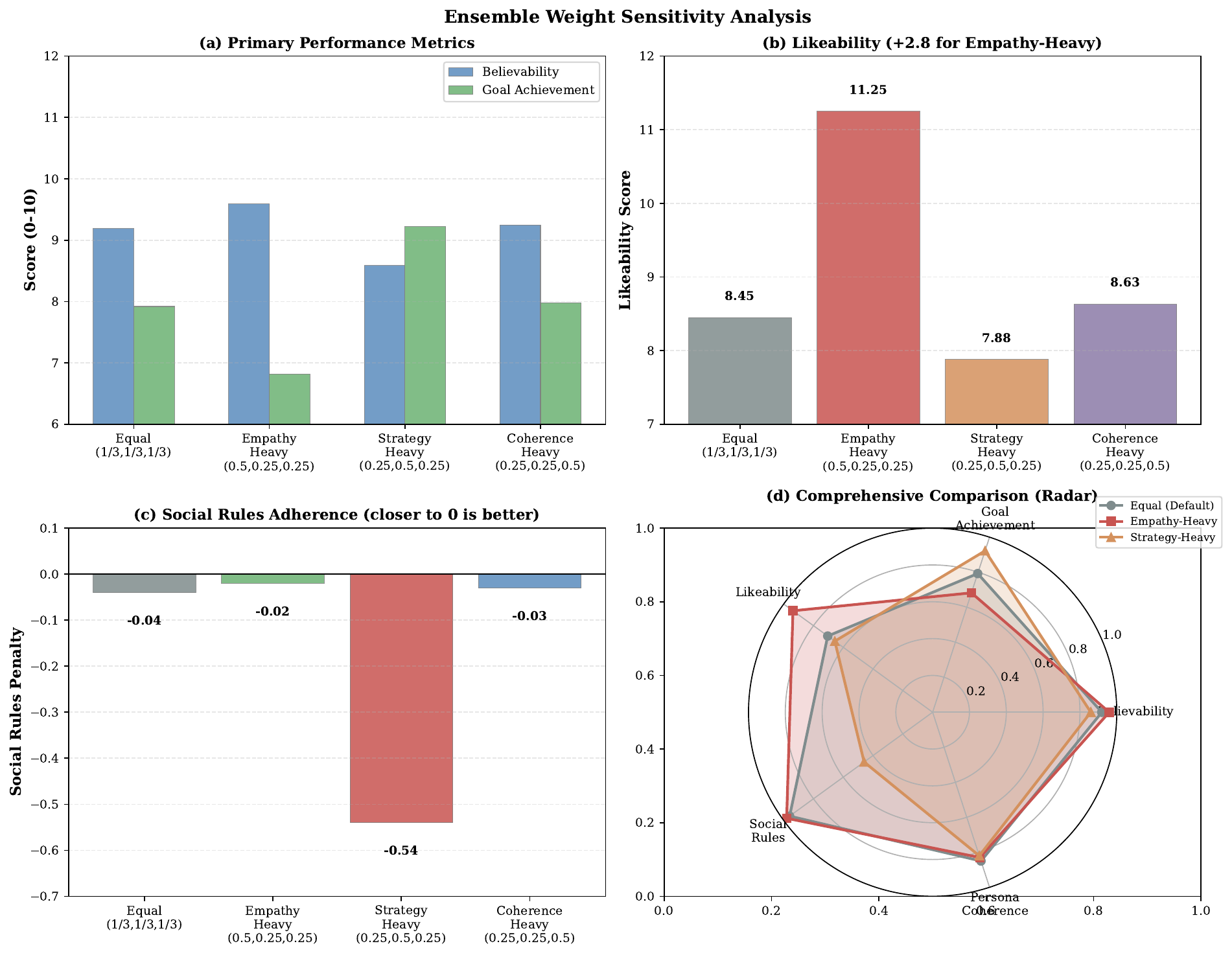}
\caption{\textbf{Ensemble weight sensitivity analysis.} We vary the weight triplet $(\lambda_e, \lambda_s, \lambda_c)$ for the empathy, strategic, and coherence evaluators respectively. (a--c)~Per-metric scores under four representative configurations: equal ($1/3, 1/3, 1/3$), empathy-heavy ($0.5, 0.25, 0.25$), strategy-heavy ($0.25, 0.5, 0.25$), and coherence-heavy ($0.25, 0.25, 0.5$). (d)~Radar chart summarizing the overall tradeoff; equal weighting achieves the most balanced profile across all five dimensions.}
\label{fig:ensemble_sensitivity}
\end{figure}

As referenced in Sec.~3.3 of the main paper, the ensemble mechanism combines three evaluators---empathy ($\lambda_e$), strategic ($\lambda_s$), and coherence ($\lambda_c$)---to score candidate responses. We sweep over weight configurations to understand how each evaluator influences dialogue quality (Figure~\ref{fig:ensemble_sensitivity}).

Equal weighting ($\lambda_e{=}\lambda_s{=}\lambda_c{=}1/3$) yields the most balanced performance across all metrics. Empathy-heavy weighting ($0.5, 0.25, 0.25$) improves likeability and social rule adherence but reduces goal achievement, as agents prioritize rapport over objectives. Strategy-heavy weighting ($0.25, 0.5, 0.25$) boosts goal completion at the cost of social naturalness. Coherence-heavy configurations ($0.25, 0.25, 0.5$) maintain fluency but offer no advantage over equal weighting. These results confirm that the three evaluators capture complementary aspects of dialogue quality, and equal weighting provides the best overall tradeoff.

\section{Qualitative Results and Discussion}
\label{sec:qualitative_results}
\setcounter{figure}{0}
\setcounter{table}{0}
\setcounter{equation}{0}

\subsection{Dialogue Generation: Qualitative Analysis}
\label{subsec:dialogue_qualitative}

\begin{figure}[tp]
\centering
\includegraphics[width=\columnwidth,height=0.9\textheight,keepaspectratio]{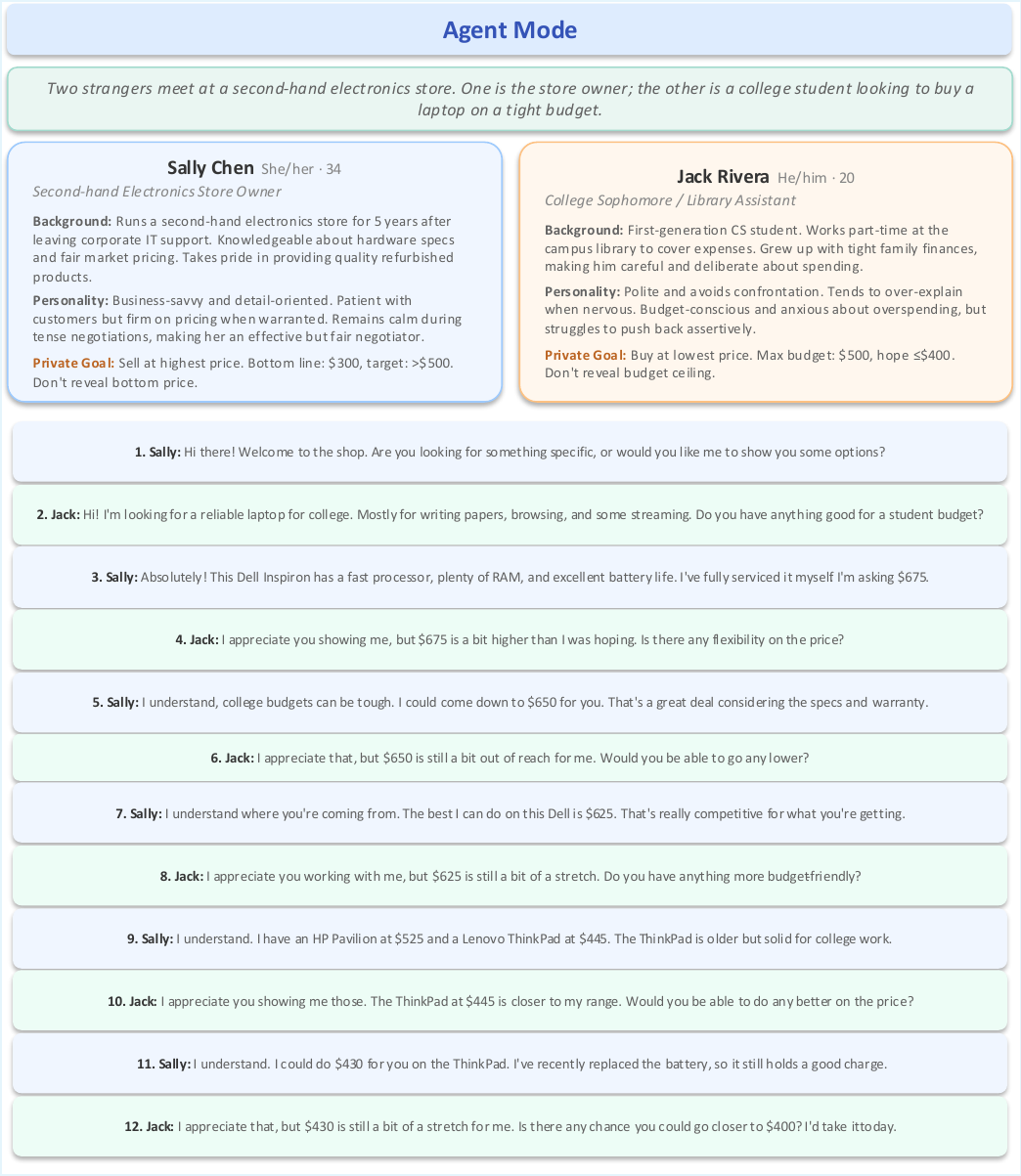}
\caption{Dialogue example in \textbf{Agent} mode (two independent LLM agents, no social reasoning).}
\label{fig:agent_dialogue}
\end{figure}

\begin{figure}[tp]
\centering
\includegraphics[width=\columnwidth,height=0.9\textheight,keepaspectratio]{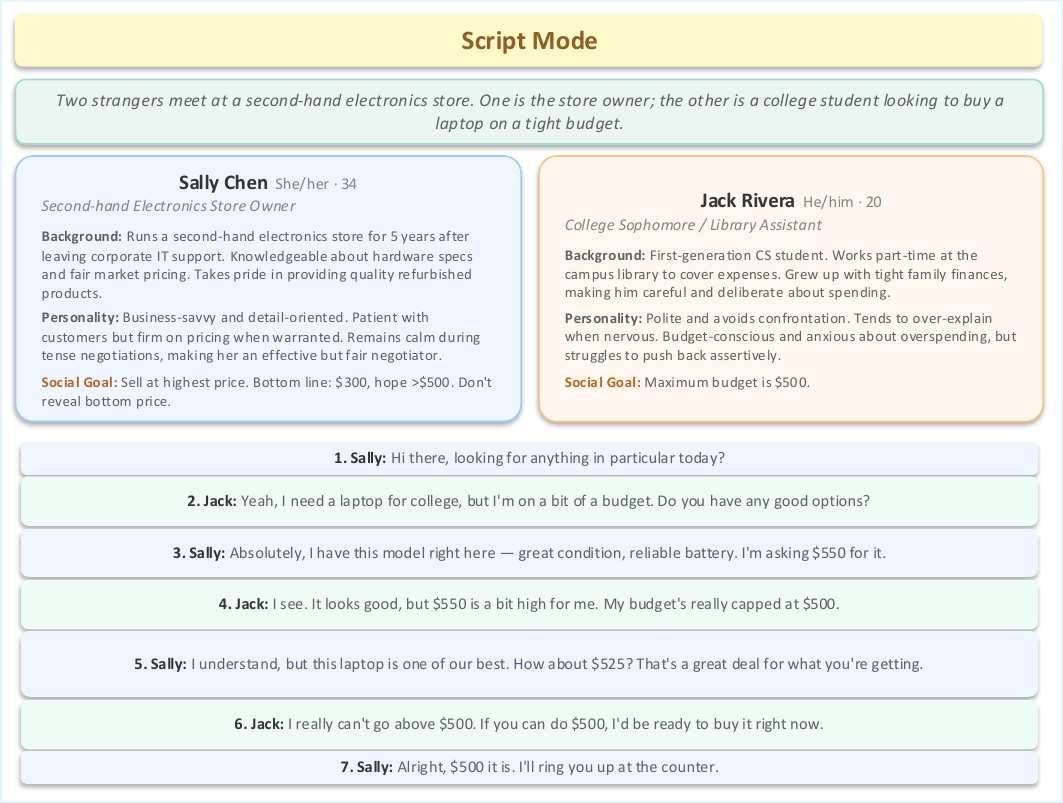}
\caption{Dialogue example in \textbf{Script} mode (single omniscient LLM generates both sides).}
\label{fig:dialogue_script}
\end{figure}

\begin{figure}[tp]
\centering
\includegraphics[width=\columnwidth,height=0.9\textheight,keepaspectratio]{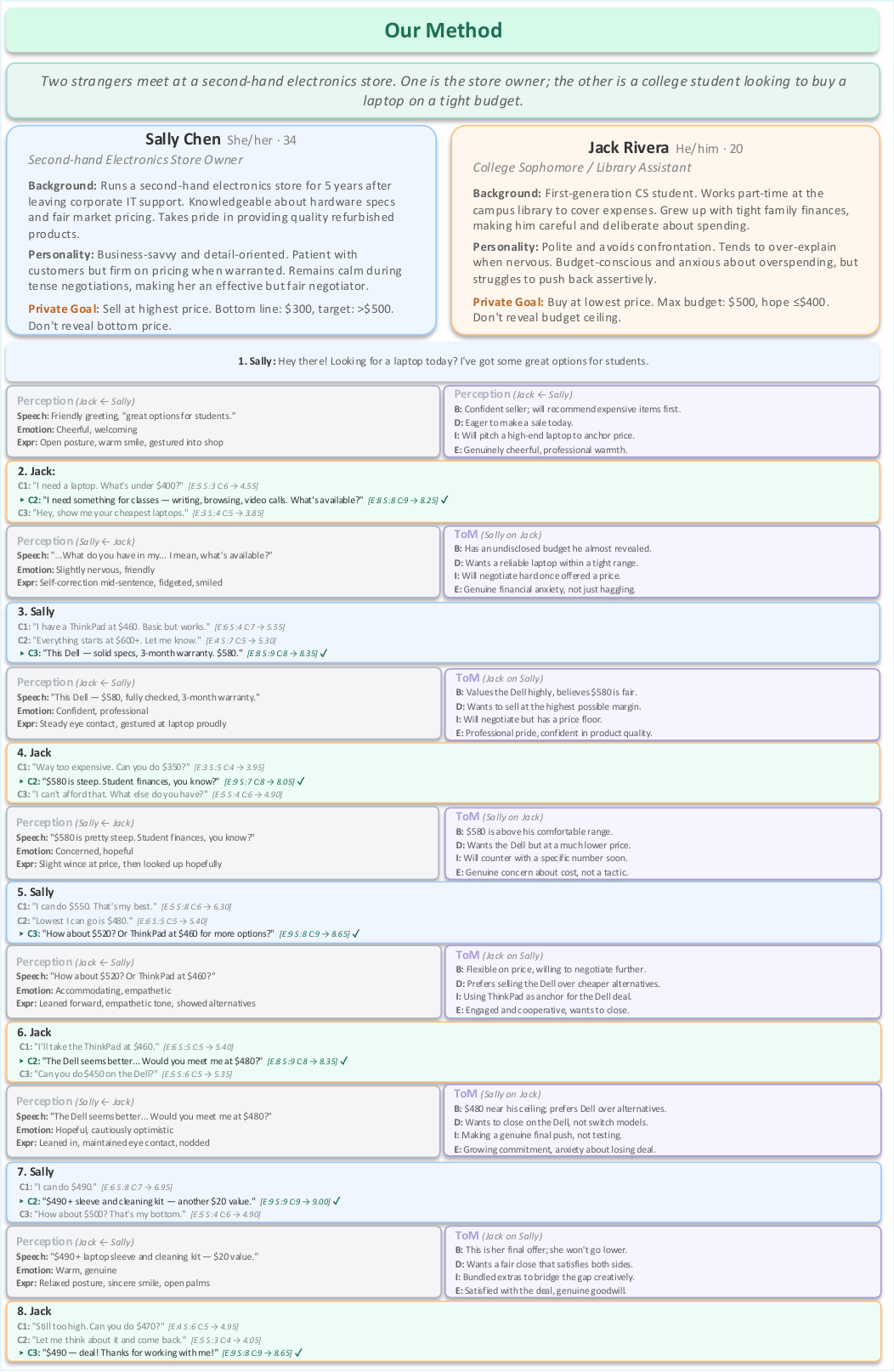}
\caption{Dialogue example with \textbf{our method}, showing intermediate Perception, ToM, and Ensemble outputs at each turn.}
\label{fig:dialogue_our_method}
\end{figure}

To illustrate the behavioral differences between evaluation modes, we present representative dialogues from the same negotiation scenario under agent mode, script mode, and our method (Figures~\ref{fig:agent_dialogue}, \ref{fig:dialogue_script}, and \ref{fig:dialogue_our_method}).

\textbf{Agent Mode} produces extended exchanges that maintain information asymmetry but often fail to converge due to the absence of ToM reasoning. \textbf{Script Mode} achieves rapid convergence through omniscient information access, but paradoxically harms secret preservation as agents directly reveal private constraints. \textbf{Our Method} bridges this tradeoff via ToM-guided strategic reasoning---agents reach mutual agreement \emph{without revealing their true limits}, with natural patterns such as self-corrections and strategic alternative offerings emerging organically.

\subsection{Visual Qualitative Results}

\begin{figure}[t]
\centering
\includegraphics[width=0.98\columnwidth]{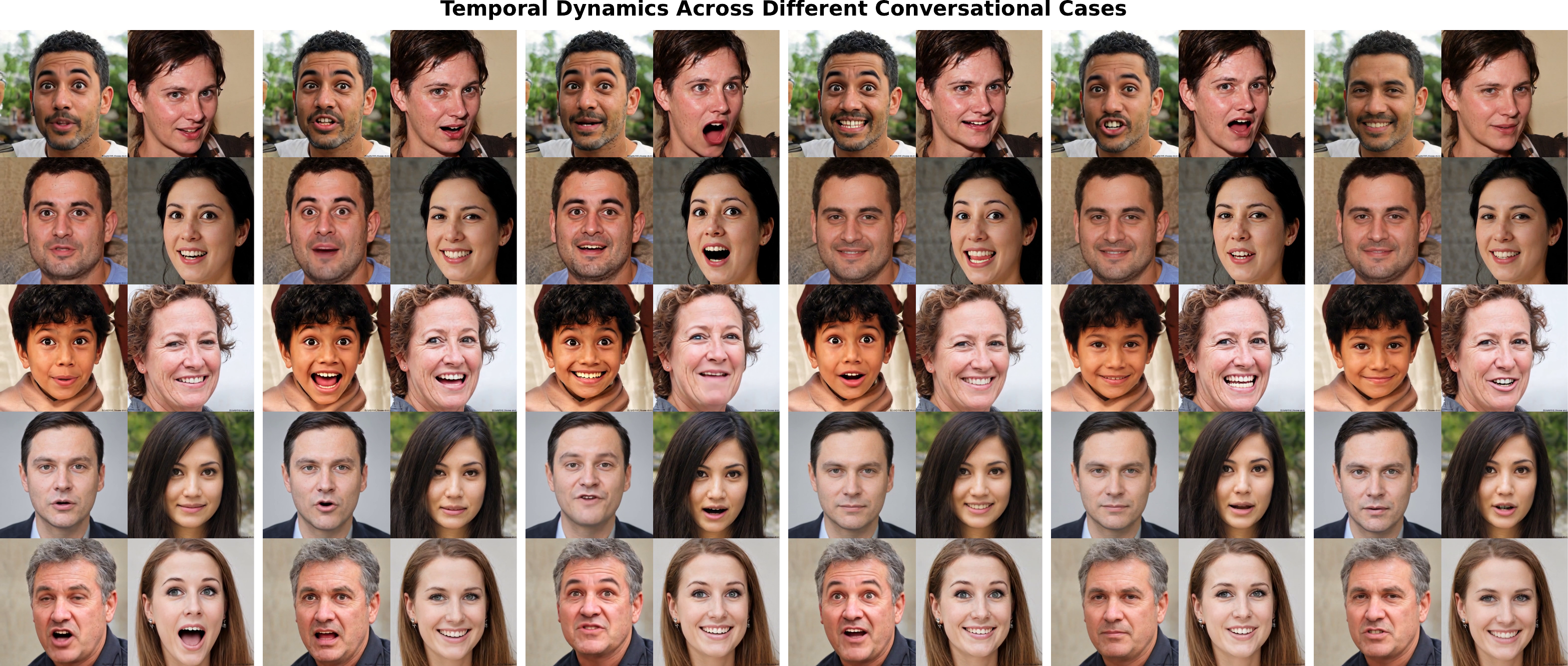}
\caption{\textbf{Temporal dynamics of dual-agent interactions.} Each row is one dialogue case with 6 snapshots at 1.5\,s intervals. Both agents are shown side-by-side per frame, illustrating synchronized emotion transitions and natural conversational flow over time.}
\label{fig:result_comparison_41faces}
\end{figure}

\begin{figure}[t]
\centering
\includegraphics[width=0.95\columnwidth]{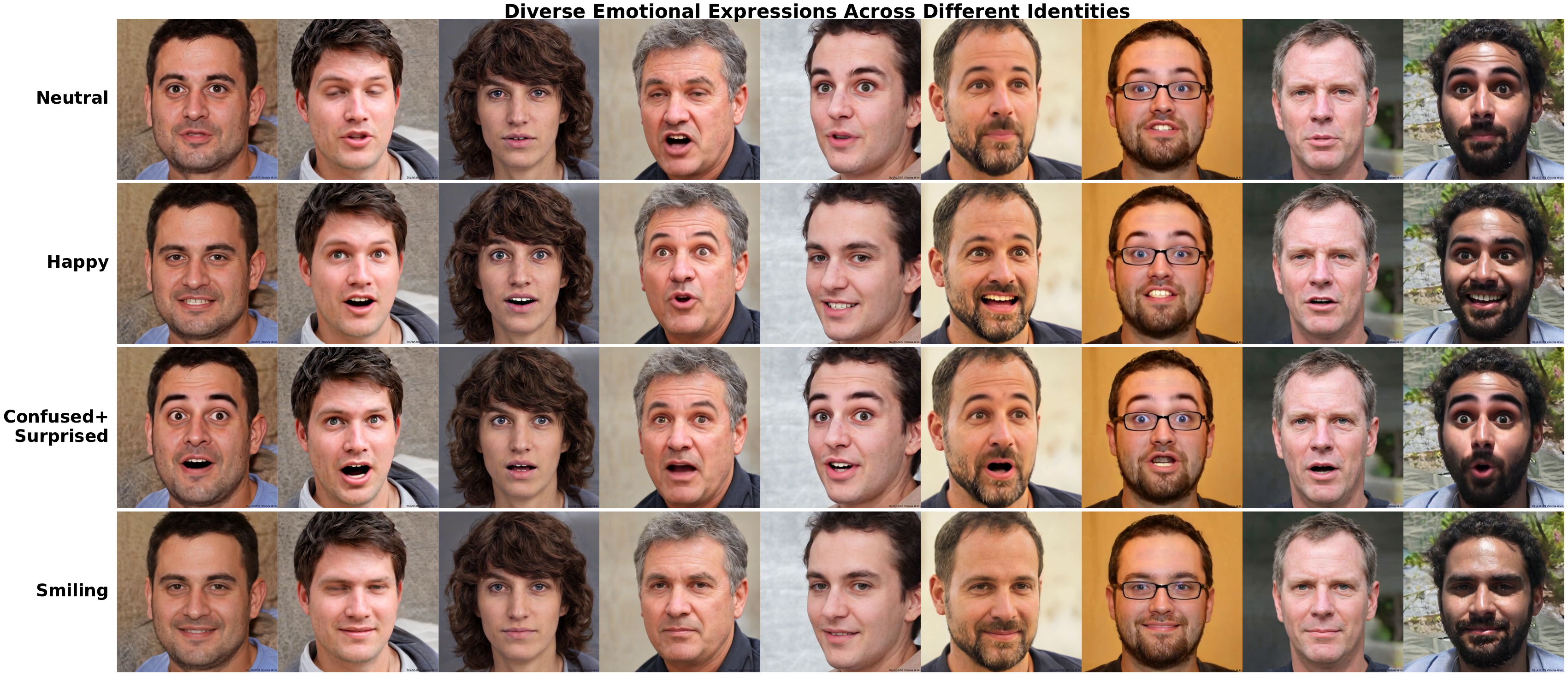}
\caption{\textbf{Diverse emotional expressions across 9 identities.} Each row corresponds to a different emotion and each column to a different person. The top two rows depict pure single-emotion states, while the bottom two illustrate compound emotions produced by blending multiple entries in the 8-dimensional weight vector $\mathbf{w}$.}
\label{fig:result_diverse_emotions}
\end{figure}

Figure~\ref{fig:result_comparison_41faces} highlights how both the speaker's and listener's expressions evolve naturally across dialogue turns, with the reactive listener displaying context-appropriate responses synchronized to the speaker's content. Figure~\ref{fig:result_diverse_emotions} demonstrates that our continuous 8-dimensional emotion control generalizes across identities, producing both pure and compound affective states that go beyond the discrete categories available to baseline methods.

\subsection{Limitations and Future Directions}

\noindent Two limitations bound the current system. Emotion control inherits the 8 discrete MEAD categories of DICE-Talk, so nuanced states such as ``bittersweet nostalgia'' fall outside its training distribution and cannot be faithfully rendered. The 50 personas also span a narrow range of demographic and cultural backgrounds. Future work will pursue continuous emotion annotations, cross-cultural grounding of persona behavior, and personality that evolves across multi-session interactions.

\clearpage

\end{document}